\pdfoutput=1
\documentclass[11pt]{article}
\usepackage{multirow}
\usepackage{multicol}

\usepackage[preprint]{acl}
\usepackage{amsmath}  
\usepackage{amssymb}  
\usepackage{times}
\usepackage{latexsym}
\usepackage[many]{tcolorbox}
\usepackage[T1]{fontenc}

\usepackage[utf8]{inputenc}

\usepackage{microtype}
\usepackage{subfigure}
\usepackage{booktabs}
\usepackage{inconsolata}

\usepackage{graphicx}

\usepackage{xcolor}
\usepackage{colortbl}
\usepackage{makecell}
\definecolor{ours-highlight}{rgb}{0.86, 0.82, 1.0}
\definecolor{grounding-data-highlight}{HTML}{ffcfdf}
\definecolor{imoveit-data-highlight}{HTML}{e0f9b5}
\definecolor{general-data-highlight}{HTML}{a5dee5}

\usepackage{tcolorbox}
\usepackage{pifont}
\newcommand{\cmark}{\ding{51}}
\newcommand{\xmark}{\ding{55}}
\newcommand{\lgcmark}{\textcolor{green}{\cmark}}
\newcommand{\lgxmark}{\textcolor{red}{\xmark}}
\usepackage{subcaption}
\usepackage{multirow}
\usepackage{afterpage}
\newcommand{\modelname}{iMOVE }
\newcommand{\datasetname}{iMOVE-IT }

\title{\raisebox{-0.01\textheight}{\includegraphics[width=0.05\textwidth]{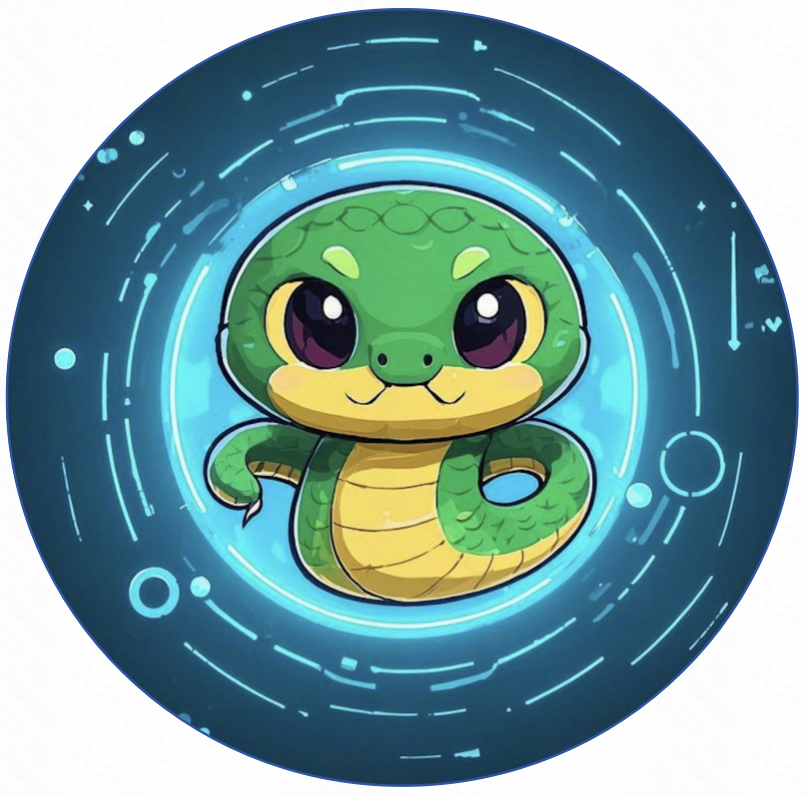}} \modelname: Instance-Motion-Aware Video Understanding}

\author{Jiaze Li\textsuperscript{1,2}, Yaya Shi\textsuperscript{1}\thanks{Corresponding author}, Zongyang Ma\textsuperscript{3}, Haoran Xu\textsuperscript{2}, Feng Cheng\textsuperscript{1} \\ \textbf{Huihui Xiao\textsuperscript{1}, Ruiwen Kang\textsuperscript{1}, Fan Yang\textsuperscript{1}, Tingting Gao\textsuperscript{1}, Di Zhang\textsuperscript{1}}  \\
  \textsuperscript{1}Kuaishou Technology  
      \textsuperscript{2}Zhejiang University \\
  \textsuperscript{3}Institute of Automation, Chinese Academy of Sciences \\
  }

\begin{document}
\maketitle
\begin{abstract}
Enhancing the fine-grained instance spatiotemporal motion perception capabilities of Video Large Language Models is crucial for improving their temporal and general video understanding. However, current models struggle to perceive detailed and complex instance motions. To address these challenges, we have made improvements from both data and model perspectives. In terms of data, we have meticulously curated \textbf{iMOVE-IT}, the first large-scale instance-motion-aware video instruction-tuning dataset. This dataset is enriched with comprehensive instance motion annotations and spatiotemporal mutual-supervision tasks, providing extensive training for the model's instance-motion-awareness. Building on this foundation, we introduce \textbf{iMOVE}, an instance-motion-aware video foundation model that utilizes Event-aware Spatiotemporal Efficient Modeling to retain informative instance spatiotemporal motion details while maintaining computational efficiency. It also incorporates Relative Spatiotemporal Position Tokens to ensure awareness of instance spatiotemporal positions. Evaluations indicate that iMOVE excels not only in video temporal understanding and general video understanding but also demonstrates significant advantages in long-term video understanding. 

\end{abstract}

\section{Introduction}

\begin{figure}
    \centering
    \includegraphics[width=0.9\linewidth]{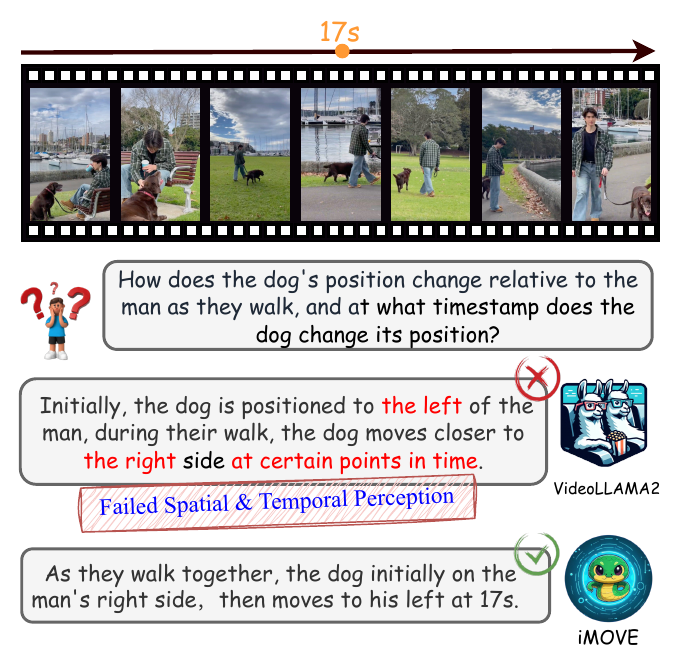}
    \caption{
     {\modelname excels at perceiving instance-level spatiotemporal motions within videos, outperforming the previous Video-LLM model, VideoLLAMA2.}
    }
    \label{fig:intro}
    \vspace{-0.5cm}
\end{figure}

Recent advancements in Video Large Language Models~\citep{video-llama, video-chatgpt, videochat, videogpt+, zhang2024llava-video}, i.e., Video-LLMs, have led to rapid development, significantly enhancing the capture of overall video semantics and achieving remarkable performance in general video understanding tasks~\citep{videomme, mvbench}. Furthermore, some works~\citep{timechat, vtimellm, lita, wang2024groundedvideollm} have begun to explore methods to enable models to comprehend detailed video semantics involving temporal information, showing progress in video temporal understanding tasks~\citep{charades-sta,activitynet}. However, current Video-LLMs struggle to accurately perceive instance spatiotemporal motions within videos, which constrains their performance in these tasks. As demonstrated in Figure \ref{fig:intro}, VideoLLAMA2~\citep{video-llama} fails to identify the dog's spatiotemporal motions.

The insufficient instance-level motion perception is primarily attributed to two aspects: data and model. On the data side, commonly used training datasets, such as ShareGPT4Video~\citep{sharegpt4video}, rely on coarse-grained video annotations and lack fine-grained instance spatiotemporal motion annotations. This deficiency makes it challenging for models to accurately perceive instance-level dynamic changes. On the model side, existing Video-LLMs~\citep{llama-vid, song2024moviechat} do not specifically consider critical spatiotemporal information during the visual token compression process, resulting in an inability to efficiently retain all instances' spatiotemporal motions in the video without loss, thereby limiting instance motion perception. Moreover, current works~\citep{timechat, lita, wang2024groundedvideollm} fail to encode both spatial and temporal position information of moving instances in the video simultaneously, leading to weak instance position awareness.

To enable exceptional capture of fine-grained instance motions and achieve superior video understanding, we first construct an  \textbf{i}nstance-\textbf{MO}tion-aware \textbf{V}id\textbf{E}o \textbf{I}nstruction-\textbf{T}uning dataset iMOVE-IT, which contains diverse, detailed and rich instance motions. iMOVE-IT defines spatiotemporal mutual-supervision goals, with tasks designed to perform spatial grounding given instance dynamic captions and time information, temporal grounding given instance dynamic captions and spatial information, and instance dynamic captioning given instance time and spatial information. These tasks work in synergy to enhance the model's instance motion awareness, thereby endowing the model with excellent temporal video understanding and general video understanding.


Building on the instance-motion-aware dataset iMOVE-IT, we further introduce iMOVE, a novel \textbf{i}nstance-\textbf{MO}tion-aware \textbf{V}id\textbf{E}o foundation model. For visual encoding, iMOVE adopts an Event-aware Spatiotemporal Efficient Modeling strategy, which adaptively segments key events in long videos while preserving the complete spatial appearances of instances and high-frame-rate temporal motions of instances within each event. This ensures the model’s detailed understanding of instance motion is not compromised by information loss. For positional encoding, iMOVE introduces Relative Spatiotemporal Position Tokens to simultaneously indicate the spatiotemporal locations of instances during motion, enabling the model to achieve high sensitivity to instance positions. These meticulous designs allow iMOVE to fully understand instance spatiotemporal motions and seamlessly integrate with the iMOVE-IT dataset, unlocking the potential of video understanding.

We conduct comprehensive evaluations on both video temporal understanding and general video understanding benchmarks. 
The results demonstrate that our approach not only achieves significant improvements in temporal understanding tasks but also excels in general video understanding tasks. 
Notably, in the zero-shot setting, iMOVE achieves a 10.5\% mIOU improvement on the temporal sentence grounding task of Charades-STA~\cite{charades-sta} and a 1.1 SODA\_c score improvement on the dense video captioning task of ActivityNet-Captions~\cite{activitynet}. 
Furthermore, in the fine-tuning setting, iMOVE surpasses classical fine-tuned expert models. Meanwhile, iMOVE not only improves general video understanding benchmarks such as MVbench and Video-MME but also significantly enhances long video understanding capabilities.
The contributions of this work are listed as follows:

\begin{itemize}
    
    \item We collect iMOVE-IT, the first large-scale instance-motion-aware video instruction-tuning dataset with rich instance motions, to improve the model’s ability to perceive spatiotemporal motions of instances.

    \item We propose iMOVE, which employs Event-aware Spatiotemporal Efficient Modeling to encode key instance spatiotemporal motions and introduces Relative Spatiotemporal Position Tokens to represent instances' spatiotemporal positions, thereby acquiring comprehensive instance motion information.

    \item iMOVE achieves outstanding instance-motion-awareness, leading to significant performance improvements in tasks related to video temporal understanding as well as general and long-term video understanding.
    
\end{itemize}

\begin{table}

\resizebox{0.47\textwidth}{!}{
\begin{tabular}{@{}ccccccc@{}}
\toprule
 & \multicolumn{2}{c}{Spatial grounding} & \multicolumn{2}{c}{Temporal Grounding} & \multicolumn{2}{c}{Instance Dynamic Captioning} \\ \cmidrule(l){2-7} 
\multirow{-2}{*}{Method} & Grasp & Generate & Grasp & Generate & Grasp & Generate \\ 
\midrule
\midrule
Elysium~\citep{wang2024elysium} & \lgxmark & \lgcmark & \lgcmark & \lgxmark & \lgcmark & \lgcmark \\
PiTe~\citep{pite} & \lgxmark & \lgcmark & \lgxmark & \lgcmark & \lgcmark & \lgcmark \\
VideoGLaMM~\citep{munasinghe2024videoglamm} & \lgxmark & \lgcmark & \lgxmark & \lgxmark & \lgcmark & \lgcmark \\
INST-IT~\citep{peng2024instit} & \lgcmark & \lgxmark & \lgcmark & \lgxmark & \lgcmark & \lgcmark \\
ViCaS~\citep{athar2024vicas} & \lgxmark & \lgcmark & \lgxmark & \lgxmark & \lgcmark & \lgcmark \\
Sa2VA~\citep{yuan2025sa2va} & \lgxmark & \lgcmark & \lgxmark & \lgxmark & \lgcmark & \lgcmark \\
VideoRefer Suite~\citep{yuan2024videorefer} & \lgcmark & \lgxmark & \lgxmark & \lgxmark & \lgcmark & \lgcmark \\
\rowcolor{ours-highlight} iMOVE-IT(Ours) & \lgcmark & \lgcmark & \lgcmark & \lgcmark & \lgcmark & \lgcmark \\
\bottomrule
\end{tabular}
}
\caption{Comparison of Instance-Motion-Aware Tasks in Different Studies.}
\label{tab:various_instance_perception_aware_tasks}
\vspace{-0.4cm}
\end{table}

\section{Related Work}
\begin{figure*}
    \centering
    \includegraphics[width=1\linewidth]{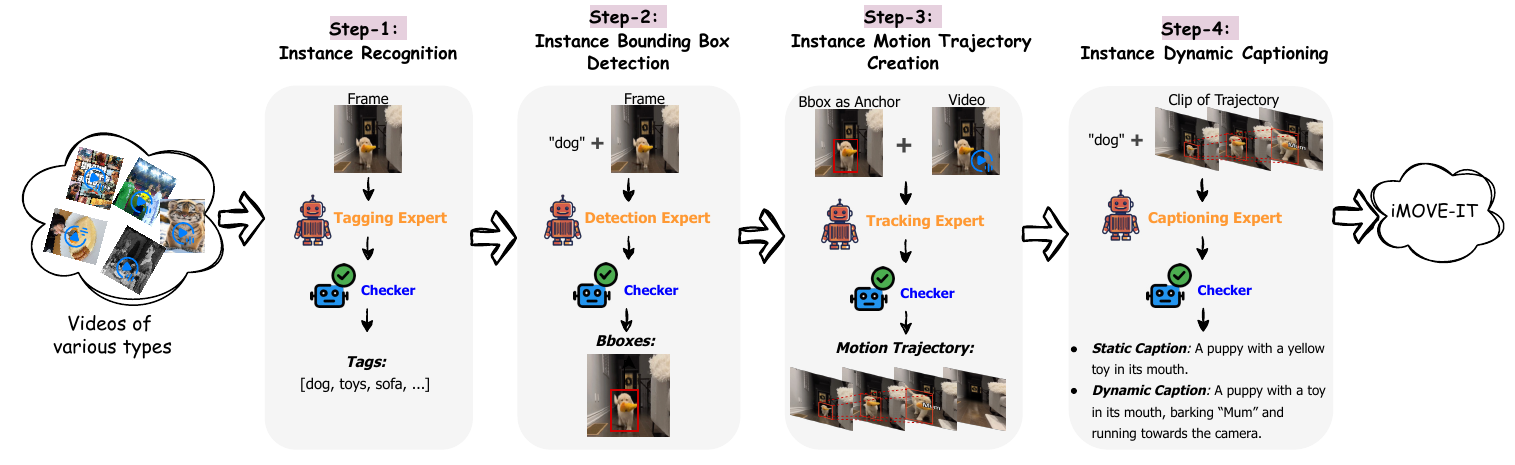}
    \caption{Instance spatiotemporal motion generation pipeline.}
    \label{fig:motion_anno}
    \vspace{-0.5cm}
\end{figure*}
\subsection{Video Large Language Models}
With the rapid advancement of large language models (LLMs), Video-LLMs have gained significant attention. Early methods, such as Video-LLAMA~\citep{video-llama} and VideoChat~\citep{videochat}, captured the overall semantics of videos through tasks like video question answering~\citep{video-chatgpt, zhang2024llava-video} and video captioning~\citep{sharegpt4video, zhang2024llava-video}. Subsequently, methods like TimeChat~\citep{timechat}, VTimeLLM~\citep{vtimellm}, and HawkEye~\citep{hawkeye} integrated tasks such as temporal grounding~\citep{charades-sta}, dense video captioning~\citep{activitynet}, and highlight detection~\citep{qvhighlights}, modeling temporal understanding. Despite these advancements, these methods still encounter challenges in fine-grained, instance-level comprehension, where accurately modeling spatiotemporal motion features is crucial for precise video understanding. Therefore, in this paper, we introduce an instance-level instruction tuning task, iMOVE-IT, and an instance-motion aware model, iMOVE.

\subsection{Instance Perception for Video Understanding}


As shown in Table~\ref{tab:various_instance_perception_aware_tasks}, recent methods have enhanced fine-grained spatiotemporal awareness through instance-level supervision tasks, categorized into spatial grounding, temporal grounding, and instance dynamic captioning, and further divided into input-side (grasp) and output-side (generate) based on model placement. Notable contributions include ElysiumTrack-1M~\citep{wang2024elysium} for single object tracking, referring tracking, and video referring expression generation; INST-IT~\citep{peng2024instit} with its instance-specific instruction tuning dataset focusing on states, transitions, and QA pairs; and VideoRefer-700K~\citep{yuan2024videorefer} providing region-level annotations with detailed captions and multi-round QA pairs for object-level video understanding. Compared to these existing methods involving tasks from either grasping or generating perspectives with limited coverage, 
our iMOVE-IT dataset provides comprehensive tasks for instance-level spatiotemporal perception.

\section{Method}
In this section, we outline the construction of iMOVE as follows: collect the dataset iMOVE-IT with an automated pipline in Sec \ref{sec:dataset} and build the model iMOVE in Sec \ref{sec:model}.

\subsection{\datasetname}
\label{sec:dataset}

\subsubsection{Instance Spatiotemporal Motion Generation}
As shown in Figure \ref{fig:motion_anno}, we propose an automatical pipeline for generating instance motion trajectories and corresponding dynamic captions. The specific steps are as follows:

\textbf{Step-1: Instance Recognition}
The initial frame of the given video is processed by the tagging expert RAM++ \citep{huang2023ram_plus} to identify all object categories present. Human checkers review these object categories and filter out some static categories, such as cloud, sky, road, and room.

\textbf{Step-2: Instance Bounding Box Detection.}

For each recognized category, we employ the open-vocabulary detection expert Grounding DINO \citep{liu2023groundingdino} to find all relevant instances and their associated bounding boxes. Subsequently, we filter out unreasonable bounding boxes, specifically those that are excessively large or those that do not match the corresponding category names based on CLIP \citep{clip} similarity scores. The filtering thresholds are optimized based on feedback from human checkers.

\textbf{Step-3: Instance Motion Trajectory Creation.} Each detected instance bounding box is treated as a tracking target and fed into the visual tracking expert SAMURAI \citep{yang2024samurai}, which outputs the instance motion trajectory. We filter out unreasonable trajectories, i.e., those with bounding boxes that differ significantly in area from the initial bounding boxes, and those with low CLIP similarity scores between the bounding boxes and their corresponding category names. The filtering thresholds are determined based on feedback from human checkers who prioritize accuracy.

\begin{figure}
    \centering
    \includegraphics[width=1.0\linewidth]{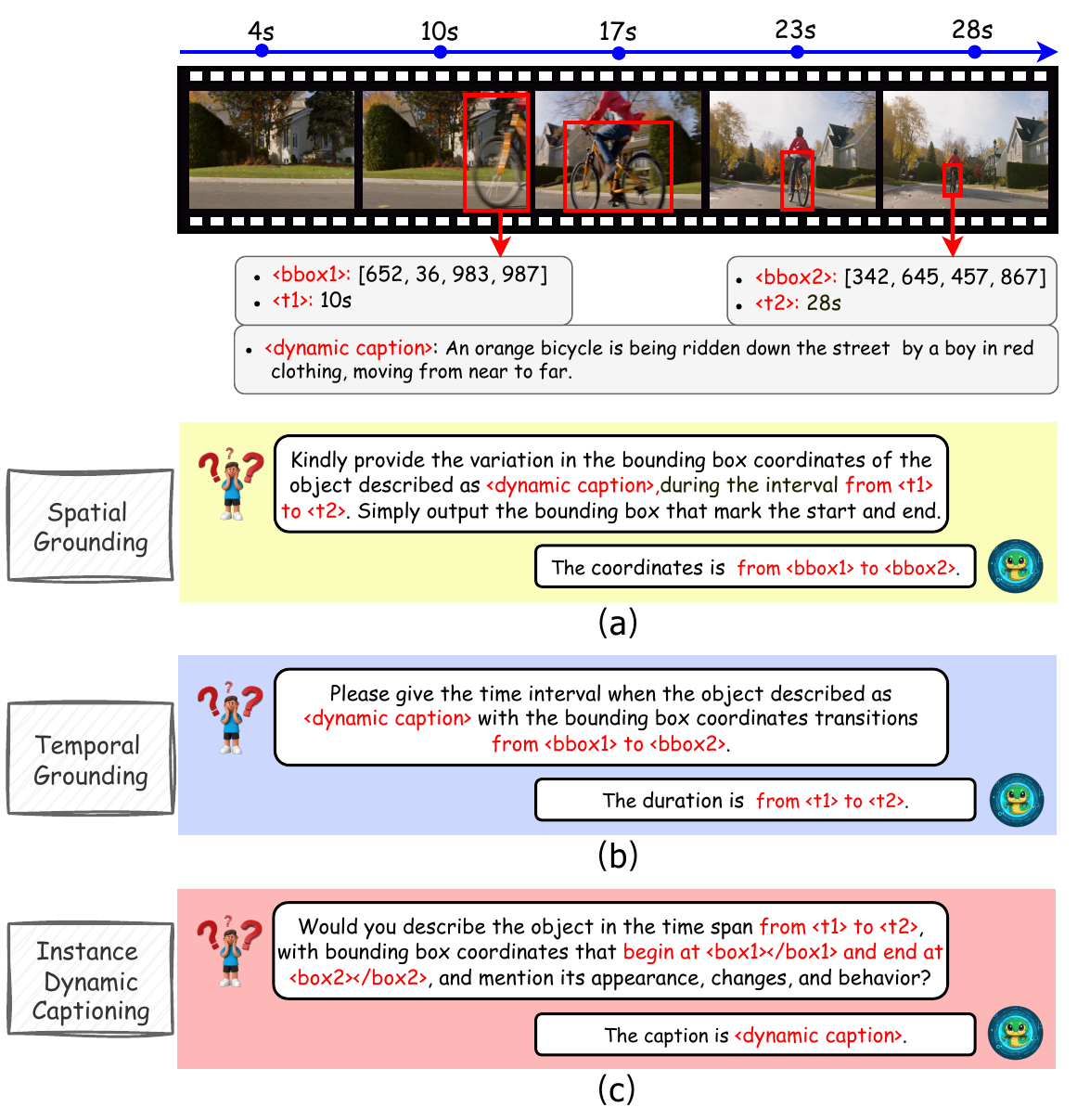}
    \caption{Examples of iMOVE-IT.}
    \label{fig:iMOVE-IT}
    \vspace{-0.5cm}
\end{figure}

\textbf{Step-4: Instance Dynamic Captioning.}
To enhance the accuracy of data generation, we query the captioning expert InternVL2-40B\citep{InternVL2024} twice, separately generating static and dynamic captions. Specifically, by combining the category name, bounding box, and motion trajectory of each instance, we first prompt InternVL2-40B to generate accurate static captions that describe the instance's appearance. Following this, based on the static captions, InternVL2-40B generates dynamic captions that describe the spatiotemporal motion of the instance. Additionally, we employ Qwen2-VL-7B-Instruct\citep{wang2024qwen2vl} as a checker to filter out erroneous dynamic captions and those that cannot uniquely refer to the instance, thereby improving the quality of the generated data.

\subsubsection{iMOVE-IT Construction}
As shown in Figure \ref{fig:iMOVE-IT}, by utilizing an automatical pipeline for data generation, we constructed iMOVE-IT, which encompasses three tasks: Spatial Grounding, Temporal Grounding, and Instance Dynamic Captioning.

\textbf{Spatial Grounding.} 
This task involves predicting the bounding box coordinates $<$bbox1, bbox2$>$ for the relevant instances in the start and end frames, based on the given dynamic caption $<$dynamic caption$>$ and the time interval $<$t1, t2$>$. This process enhances the spatial grounding capability, enabling precise localization of target instances across consecutive video frames.

\textbf{Temporal Grounding.} 
The task involves predicting the time interval $<$t1, t2$>$ corresponding to the motion trajectory of an instance based on the provided dynamic caption $<$ dynamic caption$>$ and bounding box coordinates $<$bbox1, bbox2$>$ at the start and end of the trajectory. This task enhances the model's temporal awareness of video segments, enabling it to determine time intervals corresponding to specific dynamic captions and changes in spatial positions.

\textbf{Instance Dynamic Captioning.} 
The task aims to generate a dynamic caption based on the given time interval $<$t1, t2$>$ and the bounding box coordinates $<$bbox1, bbox2$>$, which represent the start and end of the instance's motion trajectory within this time interval. This task strengthens the model's ability to match video and text, enabling it to generate dynamic caption that accurately reflect the specific instance's spatiotemporal motion within the given time interval.

The proposed iMOVE-IT dataset consists of three meticulously designed instance-level spatiotemporal mutual-supervision tasks, which provide abundant spatiotemporal supervisory signals to enhance the model's fine-grained instance spatiotemporal motion perception capabilities from the data perspective.

\subsubsection{Statistics of the iMOVE-IT Dataset}
iMOVE-IT comprises 114k video-instruction pairs from 68k unique videos, with an average duration of 109 seconds. To ensure the integrity of zero-shot setting, the dataset excludes videos from ActivityNet~\citep{activitynet} and Charades-STA~\citep{charades-sta}, while incorporating data from 11 distinct datasets. Detailed statistics are provided in Appendix~\ref{data_statics}.

\subsection{\modelname}
\label{sec:model}

\begin{figure*}
    \centering
    \includegraphics[width=1.0\linewidth]{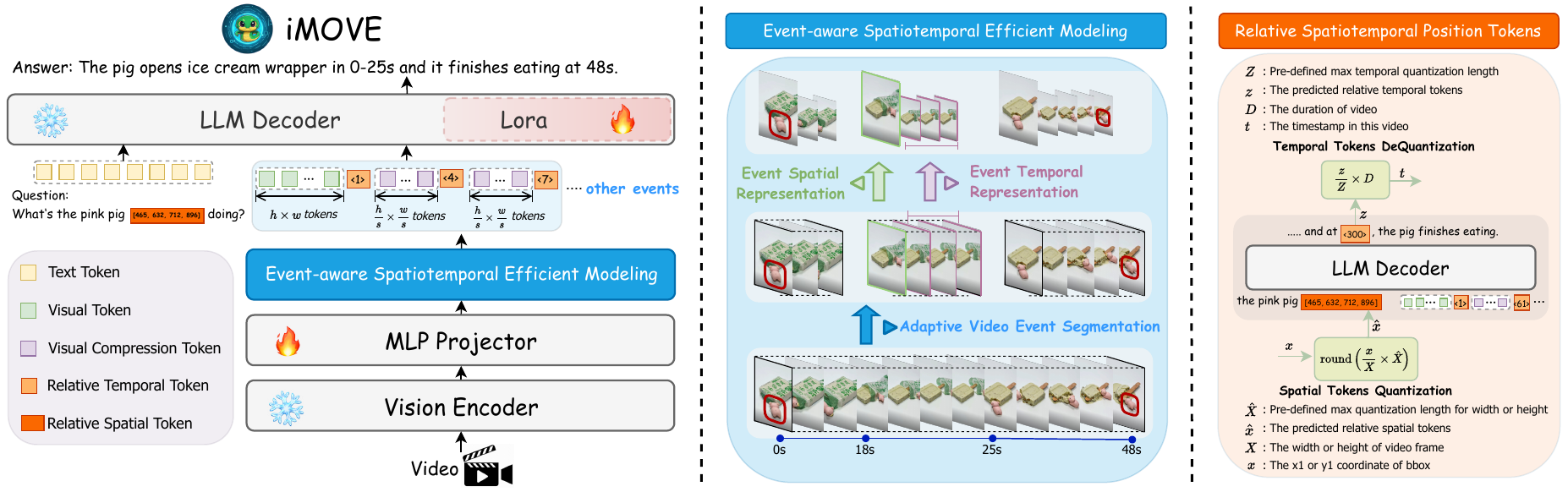}
    \caption{
    Architecture of iMOVE. iMOVE employs Event-aware Spatiotemporal Efficient Modeling to encode long videos while preserving key instance motions. Additionally, iMOVE introduces Relative Spatiotemporal Positional Token, enhancing sensitivity to spatiotemporal locations of instance motions within the video.
    }
    \label{fig:model-architecture}
    \vspace{-0.5cm}
\end{figure*}

As shown in Figure \ref{fig:model-architecture}, iMOVE adapts short Video-LLMs to perceive fine-grained spatiotemporal instance motions, thereby enhancing its temporal and general video understanding capabilities, while improving long-term video understanding.

For a long video, we sample $T$ frames and encode each into $N$ visual tokens using a visual encoder. These tokens are projected into the language semantic space, creating a feature map $\mathbf{F} \in \mathbb{R}^{\hat{h} \times \hat{w} \times d}$ for each frame, where $\hat{h}$ and $\hat{w}$ represent the dimensions of the feature map, and $d$ is the token dimension size. The complete video is represented as $\mathbf{V} = (\mathbf{F}_1, \mathbf{F}_2, \dots, \mathbf{F}_T) \in \mathbb{R}^{T \times \hat{h} \times\hat{w} \times d}$.

We further propose \textbf{Event-aware Spatiotemporal Efficient Modeling}, which reduces visual tokens in dense video encoding while preserving essential spatiotemporal motion, balancing model efficiency and video representation integrity. We also propose \textbf{Relative Spatiotemporal Position Tokens} to enhance the model's understanding of instances' spatiotemporal positions. These innovations will be presented in the subsequent sections.

\subsubsection{Event-aware Spatiotemporal Efficient Modeling}

For the raw dense visual features $\mathbf{V}$, we propose the Event-aware Spatiotemporal Efficient Modeling method, which maintains computational efficiency while preserving the complete spatial appearances of instances and the high-frame-rate temporal motions of instances. Specifically, iMOVE first adaptively segments the video into events. The frame with the most information in each event is identified as the one retaining the most spatial information. Subsequently, other frames with less information in the event undergo greater spatial compression to preserve the high-frame-rate detailed temporal and motion information of the instance. Notably, this computationally efficient process does not introduce new parameters, thereby reducing the overall model training complexity.

\textbf{Adaptive Video Event Segmentation.} We propose a novel adaptive video event segmentation method that captures event transitions in videos by identifying significant changes in inter-frame similarity. First, we calculate the cosine similarity between adjacent frame features:

\begin{equation}
S_{i, i + 1} = \frac{\mathbf{F}_i \cdot \mathbf{F}_{i+1}}{\|\mathbf{F}_i\| \|\mathbf{F}_{i+1}\|}
\end{equation}

where \( \mathbf{F}_i \) and \( \mathbf{F}_{i+1} \) are the feature vectors of adjacent frames. Then, we compute the rate of change in similarity:

\begin{equation}
d_i = |S_{i, i + 1} - S_{i - 1, i}|
\end{equation}

Next, we sort \( d_i \) in descending order and select the largest \( K - 1 \) extreme points as event boundaries, thereby segmenting the video into \( K \) events.

\textbf{Event Spatial Representation.} 
According to information entropy theory, adjacent frames with larger similarity differences carry more information, while those with smaller differences contain redundancy.
Based on this, we consider the first frame of each event, which occurs at the moment with the highest rate of information change, to be the frame with the most instance appearance information of the event and designate it as the event's spatial representation.

\begin{table*}

\centering
\small 
\resizebox{\textwidth}{!}{%
\begin{tabular}{@{}lccccccccccc@{}}
\toprule
\multicolumn{1}{l|}{\multirow{2}{*}{\textbf{Model}}} & \multicolumn{1}{c|}{\textbf{LLM}} & \multicolumn{4}{c|}{\textbf{Charades-STA}} & \multicolumn{4}{c|}{\textbf{ActivityNet-Grounding}} & \multicolumn{2}{c}{\textbf{ActivityNet-Captions}} \\ \cmidrule(l){3-12} 
\multicolumn{1}{l|}{} & \multicolumn{1}{c|}{\textbf{Scale}} & R@0.3 & R@0.5 & \multicolumn{1}{c|}{R@0.7} & \multicolumn{1}{c|}{mIoU} & R@0.3 & R@0.5 & \multicolumn{1}{c|}{R@0.7} & \multicolumn{1}{c|}{mIoU} & SODA\_c & METEOR \\ 
\midrule
\midrule
\multicolumn{12}{c}{Zero-Shot} \\ \midrule
\multicolumn{1}{l|}{Video-ChatGPT \citep{video-chatgpt}} & \multicolumn{1}{c|}{7B} & 27.2 & 6.2 & \multicolumn{1}{c|}{1.9} & \multicolumn{1}{c|}{19.7} & 19.5 & 10.6 & \multicolumn{1}{c|}{4.8} & \multicolumn{1}{c|}{14.2} & 1.9 & 2.1 \\
\multicolumn{1}{l|}{VideoChat \citep{videochat}} & \multicolumn{1}{c|}{7B} & 32.8 & 8.6 & \multicolumn{1}{c|}{0.0} & \multicolumn{1}{c|}{25.9} & 23.5 & 12.6 & \multicolumn{1}{c|}{6.0} & \multicolumn{1}{c|}{17.4} & 0.9 & 0.9 \\
\multicolumn{1}{l|}{Momentor \citep{momentor}} & \multicolumn{1}{c|}{7B} & 42.6 & 26.6 & \multicolumn{1}{c|}{11.6} & \multicolumn{1}{c|}{28.5} & \textbf{42.9} & \underline{23.0} & \multicolumn{1}{c|}{\textbf{12.4}} & \multicolumn{1}{c|}{\underline{29.3}} & \underline{2.3} & \underline{4.7} \\
\multicolumn{1}{l|}{TimeChat \citep{timechat}} & \multicolumn{1}{c|}{7B} & - & 32.2 & \multicolumn{1}{c|}{13.4} & \multicolumn{1}{c|}{-} & - & - & \multicolumn{1}{c|}{-} & \multicolumn{1}{c|}{-} & - & - \\
\multicolumn{1}{l|}{VTG-LLM \citep{vtgllm}} & \multicolumn{1}{c|}{7B} & - & 33.8 & \multicolumn{1}{c|}{15.7} & \multicolumn{1}{c|}{-} & - & - & \multicolumn{1}{c|}{-} & \multicolumn{1}{c|}{-} & - & - \\
\multicolumn{1}{l|}{HawkEye $^{\bigstar}$  \citep{hawkeye}} & \multicolumn{1}{c|}{7B} & 50.6 & 31.4 & \multicolumn{1}{c|}{14.5} & \multicolumn{1}{c|}{33.7} & \textcolor{gray}{49.1} & \textcolor{gray}{29.3} & \multicolumn{1}{c|}{\textcolor{gray}{10.7}} & \multicolumn{1}{c|}{\textcolor{gray}{32.7}} & - & - \\ 
\multicolumn{1}{l|}{PiTe$^{\bigstar}$ \citep{pite}} & \multicolumn{1}{c|}{7B} & - & - & \multicolumn{1}{c|}{-} & \multicolumn{1}{c|}{-} & \textcolor{gray}{30.4} & \textcolor{gray}{17.8} & \multicolumn{1}{c|}{\textcolor{gray}{7.8}} & \multicolumn{1}{c|}{\textcolor{gray}{22.0}} & \textcolor{gray}{5.1} & \textcolor{gray}{5.8} \\ 
\multicolumn{1}{l|}{Grounded-VideoLLM $^{\bigstar}$ \citep{wang2024groundedvideollm}} & \multicolumn{1}{c|}{4B} & 54.2 & 36.4 & \multicolumn{1}{c|}{19.7} & \multicolumn{1}{c|}{\underline{36.8}} & \textcolor{gray}{46.2} & \textcolor{gray}{30.3} & \multicolumn{1}{c|}{\textcolor{gray}{19.0}} & \multicolumn{1}{c|}{\textcolor{gray}{36.1}} & \textcolor{gray}{6.0} & \textcolor{gray}{6.8} \\ 
\multicolumn{1}{l|}{VTimeLLM $^{\bigstar}$ \citep{vtimellm}} & \multicolumn{1}{c|}{7B} & 51.0 & 27.5 & \multicolumn{1}{c|}{11.4} & \multicolumn{1}{c|}{31.2} & \textcolor{gray}{44.0} & \textcolor{gray}{27.8} & \multicolumn{1}{c|}{\textcolor{gray}{14.3}} & \multicolumn{1}{c|}{\textcolor{gray}{30.4}} & \textcolor{gray}{5.8} & \textcolor{gray}{6.8} \\
\multicolumn{1}{l|}{TIMESUITE \citep{timesuite}} & \multicolumn{1}{c|}{7B} & \underline{69.9} & \underline{48.7} & \multicolumn{1}{c|}{\underline{24.0}} & \multicolumn{1}{c|}{-} & - & - & \multicolumn{1}{c|}{-} & \multicolumn{1}{c|}{-} & - & - \\ 
\multicolumn{1}{l|}{TRACE \citep{trace}} & \multicolumn{1}{c|}{7B} & - & 40.3 & \multicolumn{1}{c|}{19.4} & \multicolumn{1}{c|}{-} & - & - & \multicolumn{1}{c|}{-} & \multicolumn{1}{c|}{-} & - & - \\ 
\multicolumn{1}{l|}{InternVL2-4B\citep{InternVL2024}} & \multicolumn{1}{c|}{4B} & 14.7 & 7.9 & \multicolumn{1}{c|}{3.3} & \multicolumn{1}{c|}{11.9} & 17.1 & 9.5 & \multicolumn{1}{c|}{4.6} & \multicolumn{1}{c|}{12.7} & 0.85 & 2.81 \\
\midrule
 \rowcolor{ours-highlight} \multicolumn{1}{l|}{iMOVE} & \multicolumn{1}{c|}{4B}  & \textbf{71.7} & \textbf{51.3} & \multicolumn{1}{c|}{\textbf{26.1}} & \multicolumn{1}{c|}{\textbf{47.3}} & \underline{42.4} & \textbf{23.1} & \multicolumn{1}{c|}{\underline{12.1}} & \multicolumn{1}{c|}{\textbf{29.7}} & \textbf{3.4} & \textbf{6.8} \\ 
 \midrule
 \midrule
\multicolumn{12}{c}{Fine-Tuning} \\ \midrule
\multicolumn{1}{l|}{Vid2Seq $^{\spadesuit}$\citep{yang2023vid2seqlargescalepretrainingvisual}} & \multicolumn{1}{c|}{-} & - & - & \multicolumn{1}{c|}{-} & \multicolumn{1}{c|}{-} & - & - & \multicolumn{1}{c|}{-} & \multicolumn{1}{c|}{-} & 5.8 & - \\
\multicolumn{1}{l|}{QD-DETR$^{\spadesuit}$\citep{moon2023querydependentvideorepresentationmoment}} & \multicolumn{1}{c|}{-} & - & 57.3 & \multicolumn{1}{c|}{32.6} & \multicolumn{1}{c|}{-} & - & - & \multicolumn{1}{c|}{-} & \multicolumn{1}{c|}{-} & - & - \\
\multicolumn{1}{l|}{UnLoc-L $^{\spadesuit}$ \citep{yan2023unlocunifiedframeworkvideo}} & \multicolumn{1}{c|}{-} & - & 60.8 & \multicolumn{1}{c|}{38.4} & \multicolumn{1}{c|}{-} & - & \underline{48.3} & \multicolumn{1}{c|}{\underline{30.2}} & \multicolumn{1}{c|}{-} & - & - \\
\multicolumn{1}{l|}{VTG-LLM \citep{vtgllm}} & \multicolumn{1}{c|}{7B} & - & 57.2 & \multicolumn{1}{c|}{33.4} & \multicolumn{1}{c|}{-} & - & - & \multicolumn{1}{c|}{-} & \multicolumn{1}{c|}{-} & - & - \\
\multicolumn{1}{l|}{HawkEye \citep{hawkeye}} & \multicolumn{1}{c|}{7B} & 72.5 & 58.3 & \multicolumn{1}{c|}{28.8} & \multicolumn{1}{c|}{\underline{49.3}} & \underline{55.9} & 34.7 & \multicolumn{1}{c|}{17.9} & \multicolumn{1}{c|}{\underline{39.1}} & - & - \\ 
\multicolumn{1}{l|}{TIMESUITE \citep{timesuite}} & \multicolumn{1}{c|}{7B} & \underline{79.4} & \underline{67.1} & \multicolumn{1}{c|}{\underline{43.0}} & \multicolumn{1}{c|}{-} & - & - & \multicolumn{1}{c|}{-} & \multicolumn{1}{c|}{-} & - & - \\ 

\multicolumn{1}{l|}{TRACE \citep{trace}} & \multicolumn{1}{c|}{7B} & - & 61.7 & \multicolumn{1}{c|}{41.4} & \multicolumn{1}{c|}{-} & - & 37.7 & \multicolumn{1}{c|}{24.0} & \multicolumn{1}{c|}{39.0} & \underline{6.0} & \underline{6.4} \\

\midrule
\rowcolor{ours-highlight} \multicolumn{1}{l|}{iMOVE+FT} & \multicolumn{1}{c|}{4B}  & \textbf{79.8} & \textbf{68.5} & \multicolumn{1}{c|}{\textbf{45.3}} & \multicolumn{1}{c|}{\textbf{57.9}} & \textbf{67.2} & \textbf{50.7} & \multicolumn{1}{c|}{\textbf{32.4}} & \multicolumn{1}{c|}{\textbf{49.3}} & \textbf{6.0} & \textbf{8.0} \\   \bottomrule
\end{tabular}%
}
\caption{Zero-Shot and Fine-Tuning results on Temporal Sentence Grounding and Dense Video Captioning tasks. \textbf{Bold} fonts highlight the best performance. \underline{Underline} highlights the second best performance. Fine-tuned expert models are marked with $^{\spadesuit}$, while non-strict zero-shot methods on ActivityNet-Captions dataset are marked with $^{\bigstar}$.}
\label{tab:results_tsg_dvc}
\vspace{-0.4cm}
\end{table*}

\textbf{Event Temporal Representation.}
Non-first frames in each event provide less spatial information than the first frame but contain rich temporal information. Therefore, we use average pooling to merge dense visual tokens spatially. Assuming the first frame feature in each event is $\mathbf{\hat{F}} \in \mathbb{R}^{{h}\times {w} \times d}$, subsequent non-first frame features are compressed based on this dimension. Using $s$ as stride, each non-first frame feature is pooled into $\mathbf{\hat{F}} \in \mathbb{R}^{\frac{h}{s} \times \frac{w}{s} \times d}$, preserving high-frame-rate temporal motions. This strategy allows the model to process more temporal frames without increasing the total visual token length, effectively modeling high-frame-rate event temporal representation.

\subsubsection{Relative Spatiotemporal Position Tokens}
Taking temporal tokens as examples, existing methods can be categorized into absolute and relative time representations. The former, such as TimeChat~\citep{timechat} and TimeMarker~\citep{chen2024timemarker}, use absolute time tokens like "2s" or "Second{8.0}". However, the broad nature of absolute time ranges and the impossibility of exhaustive enumeration make it challenging to generalize across videos of varying lengths. The latter, such as LITA~\citep{lita} and Grounded-VideoLLM~\citep{wang2024groundedvideollm}, introduce special time tokens into the tokenizer of LLMs. However, this requires modifying word embedding parameters, potentially disrupting their compatibility with the LLM's parameters and risking performance degradation, especially in small-scale fine-tuning scenarios. Therefore, in this paper, we utilize relative spatiotemporal tokens to encode bounding box coordinates and timestamp positions without adding them into the tokenizer. Unlike previous methods using relative time representations~\citep{wang2024groundedvideollm, lita}, we append a relative temporal token indicating the corresponding time to the visual tokens of each frame. These tokens are interleaved with pruned visual tokens, enabling the model to more accurately perceive the temporal information of the video. Specifically, we use quantization and dequantization on input and output sides to map actual values to relative tokens.

For temporal token, given a video of duration \( D \) seconds, we establish a bidirectional mapping between the timestamp \( t \) and a discrete value \( z \in [0, Z] \), where \( Z \) is an empirical value. The quantization process is:
\begin{equation}
z = \text{round}\left(\frac{t}{D} \times Z\right)
\end{equation}
and the dequantization process is:

\begin{equation}
t = \frac{z}{Z} \times D
\end{equation}

For spatial token, given a frame with width \( W \) and height \( H \), we establish a bidirectional mapping between the \( x \) coordinate and a discrete value \( \hat{x} \in [0, \hat{W}] \), as well as between the \( y \) coordinate and a discrete value \( \hat{y} \in [0, \hat{H}] \), in the same way as temporal token.

\begin{table*}

\centering
\small 
\resizebox{0.8\textwidth}{!}{%
\begin{tabular}{@{}l|c|cc|c|cc|c@{}}
\toprule
\multirow{2}{*}{\textbf{Model}} & \textbf{LLM} & \multicolumn{3}{c|}{\textbf{MVBench}} & \multicolumn{2}{c|}{\textbf{Video-MME(w/o subs)}} & \textbf{LongVideoBench} \\ \cmidrule(l){3-8} 
 & \textbf{Scale} & AS & AP & \multicolumn{1}{c|}{Avg} & Long & Overall & val \\ 
 \midrule
 \midrule
\multicolumn{1}{l|}{Video-LLaMA~\cite{video-llama}} & 7B & 27.5 & 25.5 & \multicolumn{1}{c|}{34.1} & - & - & - \\
\multicolumn{1}{l|}{Video-ChatGPT~\cite{video-chatgpt}} & 7B & 23.5 & 26.0 & \multicolumn{1}{c|}{32.7} & - & - & - \\
\multicolumn{1}{l|}{VideoChat2~\cite{mvbench}} & 7B & 66.0 & 47.5 & \multicolumn{1}{c|}{51.1} & 33.2 & 39.5 &  -\\
\multicolumn{1}{l|}{ST-LLM~\cite{st-llm}} & 7B & 66.0 & 53.5 & \multicolumn{1}{c|}{54.9} & 31.3 & 37.9 &  -\\
\multicolumn{1}{l|}{VideoGPT+~\cite{videogpt+}} & 7B & 69.0 & 60.0 & \multicolumn{1}{c|}{58.7} & - & - &  -\\
\multicolumn{1}{l|}{MovieChat~\cite{song2024moviechat}} & 7B & - & - & \multicolumn{1}{c|}{55.1} & 33.4 & 38.2 & - \\
\multicolumn{1}{l|}{P-LLaVA-13B~\cite{pllava}} & 13B & 66.0 & 53.0 & \multicolumn{1}{c|}{50.1} & - & - & 45.6  \\
\multicolumn{1}{l|}{LLaVA-Next-Video-34B~\cite{llava2024}} & 34B & - & - & \multicolumn{1}{c|}{-} & - & - & 50.5  \\
\multicolumn{1}{l|}{TIMESUITE \citep{timesuite}} & 7B & - & - & \multicolumn{1}{c|}{59.9} & \underline{41.9} & 46.3 & - \\
\multicolumn{1}{l|}{TRACE \citep{trace}} & 7B & - & - & \multicolumn{1}{c|}{48.1} & - & 43.8 & - \\
\multicolumn{1}{l|}{InternVL2-4B\citep{InternVL2024}} & 4B & \textbf{76.0} & \underline{62.5} & \multicolumn{1}{c|}{\underline{63.7}} & 39.2 & \underline{48.7} & \underline{50.7} \\ 
\midrule
\rowcolor{ours-highlight}\multicolumn{1}{l|}{iMOVE} & 4B & \underline{71.5} & \textbf{62.5} &  \multicolumn{1}{c|}{\textbf{63.9}} & \textbf{43.2} & \textbf{53.6} & \textbf{54.7} \\ 
\bottomrule
\end{tabular}
}
\caption{Zero-Shot results on General Video Understanding and Long-term Video Understanding tasks. \textbf{Bold} fonts highlight the best performance. \underline{Underline} highlights the second best performance.}
\label{tab:results_general_vid_understanding}
\vspace{-0.4cm}
\end{table*}

\section{Experiments}
The detailed experimental setup and hyper-parameters can be found in Appendix~\ref{experiment_setup}. Baseline and comparison details are provided in Appendix~\ref{appendix:baseline}. 
We have performed rigorous data filtering to ensure the zero-shot setting on Charades-STA and ActivityNet-Captions datasets.. Detailed training data composition and data filtering procedures are described in Appendix~\ref{data_filtering}. Qualitative analysys are available in Appendix~\ref{case_study}.

\subsection{Main Results}
To comprehensively assess the video understanding capabilities, we conducted quantitative evaluations across three task categories: Video Temporal Understanding(including Temporal Video Grounding and Dense Video Captioning), General Video Understanding and Long-term Video Understanding.
Details of the benchmarks and evaluation metrics refer to Appendix ~\ref{benchmark}. Notably, PiTe\citep{pite}, Grounded-VideoLLM\citep{wang2024groundedvideollm}, HawkEye\citep{hawkeye}, and VTimeLLM\citep{vtimellm} do not operate under a strict zero-shot setting on the ActivityNet-Captions dataset. 
Detailed explanations can be found in Appendix~\ref{data_leakage}.

\textbf{Temporal Video Grounding.} This task aims to identify the start and end timestamps of events described by a given query sentence. 
As shown in Table~\ref{tab:results_tsg_dvc}, under the zero-shot setting, iMOVE achieves mIoU accuracies of 47.3\% and 29.7\% on Charades-STA and ActivityNet-Grounding, surpassing previous SOTAs, i.e., Grounded-VideoLLM and Momentor, by margins of 10.5\% and 0.4\%. With fine-tuning, iMOVE further attains mIoU accuracies of 57.9\% and 49.3\% on these benchmarks, significantly outperforming existing SOTA methods by 10.6\% and 10.2\%.
This highlights iMOVE's superior fine-grained temporal localization capability. 

\textbf{Dense Video Captioning.} 
This task requires detecting all events in videos while providing corresponding duration time intervals and descriptions.
As observed in Table ~\ref{tab:results_tsg_dvc}, iMOVE achieves SODA\_c~\cite{fujita2020soda} and METEOR~\cite{banerjee2005meteor} scores of 3.4 and 6.8 on ActivityNet-Captions under the zero-shot setting, exceeding the prior SOTA method Momentor~\cite{momentor} with scores of 2.3 and 4.7. 
After fine-tuning, iMOVE also outperforms the specialized model TRACE~\cite{trace}.
The improvements can be attributed to iMOVE’s meticulously designed key-event retention strategy, which enables accurate capture of complete event narratives.

\textbf{General Video Understanding.}
Video-MME~\cite{videomme} and MVBench~\cite{mvbench} are used to evaluate general video understanding capabilities.
According to the results in Table~\ref{tab:results_general_vid_understanding}, despite utilizing fewer visual tokens, iMOVE obtains accuracies of 53.6 and 63.9 on Video-MME and MVBench, improving by 4.9\% and 0.2\% compared to InternVL2-4B.
While prior temporal-focused MLLMs excel at time-related tasks, they exhibit compromised performance on general video understanding tasks. 
In contrast, iMOVE performs best on both tasks, demonstrating that enhanced fine-grained instance motion modeling synergistically benefits dual capabilities.

\textbf{Long-term Video Understanding.}
The Long Video subset of Video-MME and LongVideoBench~\cite{wu2024longvideobench} serve as benchmarks for long video understanding evaluation.
The comparisons are listed in Table~\ref{tab:results_general_vid_understanding}, iMOVE acquires accuracies of 54.7 on LongVideoBench and 43.2 on Video-MME Long, both of which represent a 4\% improvement over the similarly parameter-sized Interval2-4B.
Additionally, iMOVE significantly outperforms Video-LLMs with larger parameter scales, e.g., LLaVA-Next-Video-34B~\cite{llava2024} and P-LLaVA-13B~\cite{pllava}.
This verifies that strengthening instance motion perceiving improves long video understanding.

\subsection{Ablation Study}

\begin{table}
\centering
\resizebox{\columnwidth}{!}{%
\begin{tabular}{@{}cccc|cccccc@{}}
\toprule
DVC+TAL & TSG & Geneal & iMOVE-IT & \begin{tabular}[c]{@{}c@{}}C-STA\\ mIoU\end{tabular} & \begin{tabular}[c]{@{}c@{}}ANet-G\\ mIoU\end{tabular} & \begin{tabular}[c]{@{}c@{}}ANet-Cap\\ SODA\_c\end{tabular} & \begin{tabular}[c]{@{}c@{}}LVBench\\ val\end{tabular} & \begin{tabular}[c]{@{}c@{}}V-MME\\ Overall\end{tabular} & \begin{tabular}[c]{@{}c@{}}MVB\\ Avg\end{tabular} \\ \midrule \midrule
\checkmark & & & & 25.4 & 20.3 & Fail & 46.6 & 47.0 & 57.0 \\
\checkmark & \checkmark & & & 44.8 & 27.8 & 1.8 & 50.7 & 50.3 & 57.2 \\
\checkmark & \checkmark & \checkmark & & 46.1 & 27.5 & 2.8 & 53.6 & 53.8 & 63.4 \\
\rowcolor{ours-highlight} \checkmark & \checkmark & \checkmark & \checkmark & 47.3 & 29.7 & 3.4 & 54.7 & 53.6 & 63.9 \\ \bottomrule
\end{tabular}%
}
\vspace{-0.2cm}
\caption{Ablations on training data composition. 
DVC, TAL, and TSG stands datasets for Dense Video Captioning, Temporal Action Localization and Temporal Sentence Grounding tasks. 
General includes datasets from three tasks: VideoQA, Classification, and Video Captioning. 
Fail is an inability to follow instructions.
}
\label{tab:imoveIT}
\vspace{-0.4cm}
\end{table}

\textbf{Benefits of iMOVE-IT.} 
As shown in Table \ref{tab:imoveIT}, the model performance steadily improved with the incremental inclusion of datasets from different tasks. 
The row 4 vs. row 3 comparison reveals that incorporating iMOVE-IT significantly enhances instance motion awareness, leading to improved temporal understanding, long-term video understanding, and general video understanding capabilities.

\begin{table}
\centering

\vspace{-0.2cm}
\resizebox{0.95\columnwidth}{!}{%
\begin{tabular}{@{}c|ccc|ccc|cc@{}}
\toprule
Row & Rand & Uniform & Event & K        & s & \# of tokens & \begin{tabular}[c]{@{}c@{}}C-STA\\ mIoU\end{tabular} & \begin{tabular}[c]{@{}c@{}}V-MME\\ Overall\end{tabular} \\ \midrule \midrule
1    &  \checkmark    &         &       &      24    & 2   &  2688            &   43.2    &    50.0   \\
2    &      &      \checkmark   &       &      24    & 2  &   2688       &    43.6   &  49.9     \\
\rowcolor{ours-highlight}3    &      &         &    \checkmark   & 24 & 2 & 2688         & 44.8  & 50.3  \\ \midrule
4    &      &         &    \checkmark   & 0        & 2 & 1536         & 43.8  & 49.2  \\
5    &      &         &   \checkmark    & 12       & 2 & 2112         & 45.0  & 49.8  \\
6    &      &         &   \checkmark    & 48       & 2 & 3840         & 45.1  & 51.3  \\
7    &      &         &   \checkmark    & 96       & 2 & 6144         & 46.2  & 51.1  \\ \midrule
8    &      &         &   \checkmark    & 24       & 1 & 6144         & 46.2  & 51.1  \\
9    &      &         &   \checkmark    & 24       & 4 & 1824         & 44.9  & 49.9  \\
10   &      &         &   \checkmark    & 24       & 8 & 1608         & 44.2  & 49.1  \\ \bottomrule
\end{tabular}%
}
\vspace{-0.2cm}
\caption{Effect of Event-aware Spatiotemporal Efficient Modeling. Rand denotes randomly selecting the first frame of the Event, Uniform denotes uniformly selecting frames, and Event denotes our method.}
\label{tab:about_K_r}
\vspace{-0.1cm}
\end{table}

\textbf{Effectiveness of Event-aware Spatiotemporal Efficient Modeling.} 
The first three rows of Table \ref{tab:about_K_r} demonstrate that our proposed Event-aware Spatiotemporal Efficient Modeling outperforms both random and uniform frame selection strategies. Furthermore, as observed from rows 3 to 5, increasing the number of segmented events \(K\) leads to continuous improvements on Charades-STA and Video-MME datasets. 
Rows 8 to 10 indicate that as the non-first frame pooling rate \(s\) increases, there is a loss of spatial information, which results in a decline in the model's temporal understanding and generalization capabilities. Conversely, when the pooling rate is too small (\(s = 1\)), the excessive number of tokens adversely affects the model's training and inference efficiency. Balancing information loss and efficiency, we selected a pooling rate of (\(K = 24\)) and (\(s = 2\)) as the optimal trade-off. The composition of the data for this ablation study can be found in Appendix ~\ref{dataset_composition_abl}.


\begin{table}
\centering

\vspace{-0.2cm}
\resizebox{0.95\columnwidth}{!}{%
\begin{tabular}{@{}c|cc|c|c|c|cc@{}}
\toprule
\multirow{2}{*}{Row} & \multirow{2}{*}{RT} & \multirow{2}{*}{AT} & \multirow{2}{*}{Add-VT} & \multirow{2}{*}{Add-Tokenizer} & \multirow{2}{*}{RST} & \multirow{2}{*}{\begin{tabular}[c]{@{}c@{}}C-STA\\ mIoU\end{tabular}} & \multirow{2}{*}{\begin{tabular}[c]{@{}c@{}}ANet-G\\ mIoU\end{tabular}} \\
                      &                     &                     &                         &                                &                      &                                                                       &                                                                        \\ \midrule \midrule
1                     &            \checkmark         &                     &                         &                                &                      & 44.9                                                                  & 26.3                                                                   \\
2                     &          \checkmark           &                     &        \checkmark                 &                                &                      & 46.6                                                                  & 29.2                                                                   \\
\rowcolor{ours-highlight} 3                     &         \checkmark            &                     &        \checkmark                 &                                &               \checkmark       & \textbf{47.3}                                                         & \textbf{29.7}                                                          \\
4                     &                     &        \checkmark             &         \checkmark                &                                &            \checkmark          & 45.7                                                                  & 28.4                                                                   \\
5                     &           \checkmark          &                     &         \checkmark                &                            \checkmark    &           \checkmark           & 44.4                                                                  & 27.6                                                                   \\ \bottomrule
\end{tabular}%
}
\vspace{-0.2cm}
\caption{Effect of Relative Spatiotemporal Position Tokens: RT and AT denote relative and absolute temporal representations. Add-VT appends the temporal token after each frame's visual tokens, while Add-Tokenizer incorporates temporal position tokens into the tokenizer. RST represents Relative Spatial Position Token.}
\label{tab:about_temporal_tokens}
\vspace{-0.4cm}
\end{table}

\textbf{About of Relative Spatiotemporal Position Tokens.} Table~\ref{tab:about_temporal_tokens} shows that row 2 and row1 indicate that adding corresponding temporal token after each frame's visual token is beneficial for achieving better model performance. The performance improvement of row 3 compared to row 4 demonstrates that relative temporal representation outperforms absolute temporal representation in our experiments. The performance decline of row 5 relative to row 3 suggests that introducing additional temporal markers in the LLM's tokenizer results in suboptimal outcomes. Meanwhile, the performance improvement of row 3 compared to row 2 proves the effectiveness of the Relative Spatial Position Token. In summary, these findings demonstrate the effectiveness of the Relative Spatiotemporal Position Tokens.

\begin{table}
\centering

\vspace{-0.2cm}
\begin{center}
\resizebox{0.9\columnwidth}{!}{%
\begin{tabular}{@{}c|c|c|cc|c@{}}
\toprule
\multirow{2}{*}{Architecture} & C-STA & ANet-G & \multicolumn{2}{c|}{ANet-Cap} & LVBench \\ \cmidrule(l){2-6} 
                              & mIoU  & mIoU   & SODA\_c        & METEOR       & val     \\ \midrule \midrule
Phi3.5-V        & 45.5  & 29.2   & 3.3            & 7.2          & 53.4    \\ \midrule
\rowcolor{ours-highlight}InternVL2-4B                     & 47.3  & 29.7   & 3.4            & 6.8          & 54.7    \\ \bottomrule
\end{tabular}%
}
\end{center}
\vspace{-0.2cm}
\caption{Generalization Study. Phi3.5-V denotes Phi3.5-Vision-Instruct-3.8B.}
\label{tab:general}
\vspace{-0.4cm}
\end{table}

\textbf{Generalization Study}. To validate the generalization capability of the proposed method, we employ an image MLLM, Phi3.5-Vision-Instruct-3.8B~\citep{phi3v}, as our base MLLM. As shown in Table~\ref{tab:general}, using an image MLLM as the base achieves comparable performance to a short VideoLLM. Notably, it improves the METEOR score by 0.4 points on ActivityNet-Captions. This indicates that our model architecture and dataset not only enable short Video-LLMs to adapt to perceiving fine-grained spatiotemporal instance motions but are also applicable to image MLLMs.

\section{Conclusion}

In this paper, we enhanced the fine-grained instance spatiotemporal motion perception in Video-LLMs by making improvements from both data and model perspectives, thereby boosting their capabilities in temporal and general video understanding. Data-wise, we propose \textbf{iMOVE-IT}, a large instance-motion-aware video dataset with spatiotemporal mutual-supervision tasks to enhance instance spatiotemporal motion learning. Model-wise, we develop \textbf{iMOVE} featuring Event-aware Spatiotemporal Efficient Modeling for efficiently preserving motion details, and Relative Spatiotemporal Position Tokens for accurate spatial-temporal positioning.

Evaluations indicate that iMOVE excels in video temporal understanding and general video understanding, with significant advantages in long-term video understanding, offering valuable insights for video understanding research.

\bibliography{custom}

\begin{thebibliography}{53}
\providecommand{\natexlab}[1]{#1}

\bibitem[{Abdin et~al.(2024)Abdin, Aneja, Awadalla, Awadallah, Awan, Bach, Bahree, Bakhtiari, Bao, Behl et~al.}]{abdin2024phi}
Marah Abdin, Jyoti Aneja, Hany Awadalla, Ahmed Awadallah, Ammar~Ahmad Awan, Nguyen Bach, Amit Bahree, Arash Bakhtiari, Jianmin Bao, Harkirat Behl, et~al. 2024.
\newblock Phi-3 technical report: A highly capable language model locally on your phone.
\newblock \emph{arXiv preprint arXiv:2404.14219}.

\bibitem[{Athar et~al.(2024)Athar, Deng, and Chen}]{athar2024vicas}
Ali Athar, Xueqing Deng, and Liang-Chieh Chen. 2024.
\newblock Vicas: A dataset for combining holistic and pixel-level video understanding using captions with grounded segmentation.
\newblock \emph{arXiv preprint arXiv: 2412.09754}.

\bibitem[{Banerjee and Lavie(2005)}]{banerjee2005meteor}
Satanjeev Banerjee and Alon Lavie. 2005.
\newblock Meteor: An automatic metric for mt evaluation with improved correlation with human judgments.
\newblock In \emph{Proceedings of the acl workshop on intrinsic and extrinsic evaluation measures for machine translation and/or summarization}, pages 65--72.

\bibitem[{Caba~Heilbron et~al.(2015)Caba~Heilbron, Escorcia, Ghanem, and Carlos~Niebles}]{activitynet}
Fabian Caba~Heilbron, Victor Escorcia, Bernard Ghanem, and Juan Carlos~Niebles. 2015.
\newblock Activitynet: A large-scale video benchmark for human activity understanding.
\newblock In \emph{CVPR}, pages 961--970.

\bibitem[{Chen et~al.(2024{\natexlab{a}})Chen, Wei, Li, Dong, Zhang, Zang, Chen, Duan, Lin, Tang, Yuan, Qiao, Lin, Zhao, and Wang}]{sharegpt4video}
Lin Chen, Xilin Wei, Jinsong Li, Xiaoyi Dong, Pan Zhang, Yuhang Zang, Zehui Chen, Haodong Duan, Bin Lin, Zhenyu Tang, Li~Yuan, Yu~Qiao, Dahua Lin, Feng Zhao, and Jiaqi Wang. 2024{\natexlab{a}}.
\newblock Sharegpt4video: Improving video understanding and generation with better captions.
\newblock \emph{arXiv preprint arXiv:2406.04325}.

\bibitem[{Chen et~al.(2024{\natexlab{b}})Chen, Lan, Yuan, Jie, and Ma}]{chen2024timemarker}
Shimin Chen, Xiaohan Lan, Yitian Yuan, Zequn Jie, and Lin Ma. 2024{\natexlab{b}}.
\newblock Timemarker: A versatile video-llm for long and short video understanding with superior temporal localization ability.
\newblock \emph{arXiv preprint arXiv: 2411.18211}.

\bibitem[{Chen et~al.(2024{\natexlab{c}})Chen, Wang, Tian, Ye, Gao, Cui, Tong, Hu, Luo, Ma et~al.}]{InternVL2024}
Zhe Chen, Weiyun Wang, Hao Tian, Shenglong Ye, Zhangwei Gao, Erfei Cui, Wenwen Tong, Kongzhi Hu, Jiapeng Luo, Zheng Ma, et~al. 2024{\natexlab{c}}.
\newblock Internvl2: Better than the best—expanding performance boundaries of open-source multimodal models with the progressive scaling strategy.
\newblock \url{https://internvl.github.io/blog/2024-07-02-InternVL-2.0/}.

\bibitem[{Chen et~al.(2024{\natexlab{d}})Chen, Wu, Wang, Su, Chen, Xing, Zhong, Zhang, Zhu, Lu et~al.}]{chen2024internvl}
Zhe Chen, Jiannan Wu, Wenhai Wang, Weijie Su, Guo Chen, Sen Xing, Muyan Zhong, Qinglong Zhang, Xizhou Zhu, Lewei Lu, et~al. 2024{\natexlab{d}}.
\newblock Internvl: Scaling up vision foundation models and aligning for generic visual-linguistic tasks.
\newblock In \emph{Proceedings of the IEEE/CVF Conference on Computer Vision and Pattern Recognition}, pages 24185--24198.

\bibitem[{Fu et~al.(2024)Fu, Dai, Luo, Li, Ren, Zhang, Wang, Zhou, Shen, Zhang et~al.}]{videomme}
Chaoyou Fu, Yuhan Dai, Yondong Luo, Lei Li, Shuhuai Ren, Renrui Zhang, Zihan Wang, Chenyu Zhou, Yunhang Shen, Mengdan Zhang, et~al. 2024.
\newblock Video-mme: The first-ever comprehensive evaluation benchmark of multi-modal llms in video analysis.
\newblock \emph{arXiv preprint arXiv:2405.21075}.

\bibitem[{Fujita et~al.(2020)Fujita, Hirao, Kamigaito, Okumura, and Nagata}]{fujita2020soda}
Soichiro Fujita, Tsutomu Hirao, Hidetaka Kamigaito, Manabu Okumura, and Masaaki Nagata. 2020.
\newblock Soda: Story oriented dense video captioning evaluation framework.
\newblock In \emph{Computer Vision--ECCV 2020: 16th European Conference, Glasgow, UK, August 23--28, 2020, Proceedings, Part VI 16}, pages 517--531. Springer.

\bibitem[{Gao et~al.(2017)Gao, Sun, Yang, and Nevatia}]{charades-sta}
Jiyang Gao, Chen Sun, Zhenheng Yang, and Ram Nevatia. 2017.
\newblock Tall: Temporal activity localization via language query.
\newblock In \emph{ICCV}, pages 5267--5275.

\bibitem[{Guo et~al.(2024{\natexlab{a}})Guo, Liu, Li, Tang, Chen, and Zhao}]{vtgllm}
Yongxin Guo, Jingyu Liu, Mingda Li, Xiaoying Tang, Xi~Chen, and Bo~Zhao. 2024{\natexlab{a}}.
\newblock Vtg-llm: Integrating timestamp knowledge into video llms for enhanced video temporal grounding.
\newblock \emph{arXiv preprint arXiv:2405.13382}.

\bibitem[{Guo et~al.(2024{\natexlab{b}})Guo, Liu, Li, Tang, Liu, and Chen}]{trace}
Yongxin Guo, Jingyu Liu, Mingda Li, Xiaoying Tang, Qingbin Liu, and Xi~Chen. 2024{\natexlab{b}}.
\newblock Trace: Temporal grounding video llm via causal event modeling.
\newblock \emph{arXiv preprint arXiv: 2410.05643}.

\bibitem[{Hu et~al.(2022)Hu, Shen, Wallis, Allen-Zhu, Li, Wang, Wang, and Chen}]{lora}
Edward~J Hu, Yelong Shen, Phillip Wallis, Zeyuan Allen-Zhu, Yuanzhi Li, Shean Wang, Lu~Wang, and Weizhu Chen. 2022.
\newblock Lora: Low-rank adaptation of large language models.
\newblock In \emph{ICLR}.

\bibitem[{Huang et~al.(2024{\natexlab{a}})Huang, Wang, Chen, Song, and Zhu}]{vtimellm}
Bin Huang, Xin Wang, Hong Chen, Zihan Song, and Wenwu Zhu. 2024{\natexlab{a}}.
\newblock Vtimellm: Empower llm to grasp video moments.
\newblock In \emph{CVPR}.

\bibitem[{Huang et~al.(2024{\natexlab{b}})Huang, Liao, Radhakrishnan, Yin, Molchanov, Yu, and Kautz}]{lita}
De-An Huang, Shijia Liao, Subhashree Radhakrishnan, Hongxu Yin, Pavlo Molchanov, Zhiding Yu, and Jan Kautz. 2024{\natexlab{b}}.
\newblock Lita: Language instructed temporal-localization assistant.
\newblock \emph{arXiv preprint arXiv:2403.19046}.

\bibitem[{Huang et~al.(2023)Huang, Huang, Zhang, Tian, Feng, Zhang, Xie, Li, and Zhang}]{huang2023ram_plus}
Xinyu Huang, Yi-Jie Huang, Youcai Zhang, Weiwei Tian, Rui Feng, Yuejie Zhang, Yanchun Xie, Yaqian Li, and Lei Zhang. 2023.
\newblock Open-set image tagging with multi-grained text supervision.
\newblock \emph{arXiv e-prints}, pages arXiv--2310.

\bibitem[{Krishna et~al.(2017)Krishna, Hata, Ren, Fei-Fei, and Carlos~Niebles}]{krishna2017dense}
Ranjay Krishna, Kenji Hata, Frederic Ren, Li~Fei-Fei, and Juan Carlos~Niebles. 2017.
\newblock Dense-captioning events in videos.
\newblock In \emph{Proceedings of the IEEE international conference on computer vision}, pages 706--715.

\bibitem[{Lei et~al.(2021)Lei, Berg, and Bansal}]{qvhighlights}
Jie Lei, Tamara~L Berg, and Mohit Bansal. 2021.
\newblock Detecting moments and highlights in videos via natural language queries.
\newblock In \emph{NeurIPS}, pages 11846--11858.

\bibitem[{Li et~al.(2023{\natexlab{a}})Li, He, Wang, Li, Wang, Luo, Wang, Wang, and Qiao}]{videochat}
KunChang Li, Yinan He, Yi~Wang, Yizhuo Li, Wenhai Wang, Ping Luo, Yali Wang, Limin Wang, and Yu~Qiao. 2023{\natexlab{a}}.
\newblock Videochat: Chat-centric video understanding.
\newblock \emph{arXiv preprint arXiv:2305.06355}.

\bibitem[{Li et~al.(2024)Li, Wang, He, Li, Wang, Liu, Wang, Xu, Chen, Luo et~al.}]{mvbench}
Kunchang Li, Yali Wang, Yinan He, Yizhuo Li, Yi~Wang, Yi~Liu, Zun Wang, Jilan Xu, Guo Chen, Ping Luo, et~al. 2024.
\newblock Mvbench: A comprehensive multi-modal video understanding benchmark.
\newblock In \emph{CVPR}.

\bibitem[{Li et~al.(2023{\natexlab{b}})Li, Wang, and Jia}]{llama-vid}
Yanwei Li, Chengyao Wang, and Jiaya Jia. 2023{\natexlab{b}}.
\newblock Llama-vid: An image is worth 2 tokens in large language models.
\newblock \emph{arXiv preprint arXiv:2311.17043}.

\bibitem[{Liu et~al.(2024{\natexlab{a}})Liu, Li, Tang, Ge, Shan, and Li}]{st-llm}
Ruyang Liu, Chen Li, Haoran Tang, Yixiao Ge, Ying Shan, and Ge~Li. 2024{\natexlab{a}}.
\newblock St-llm: Large language models are effective temporal learners.
\newblock In \emph{ECCV}.

\bibitem[{Liu et~al.(2024{\natexlab{b}})Liu, Zeng, Ren, Li, Zhang, Yang, Jiang, Li, Yang, Su, Zhu, and Zhang}]{liu2023groundingdino}
Shilong Liu, Zhaoyang Zeng, Tianhe Ren, Feng Li, Hao Zhang, Jie Yang, Qing Jiang, Chunyuan Li, Jianwei Yang, Hang Su, Jun Zhu, and Lei Zhang. 2024{\natexlab{b}}.
\newblock Grounding {DINO:} marrying {DINO} with grounded pre-training for open-set object detection.
\newblock In \emph{ECCV}, pages 38--55.

\bibitem[{Liu et~al.(2024{\natexlab{c}})Liu, Ding, Huang, Zhang, Zhao, and Wang}]{pite}
Yang Liu, Pengxiang Ding, Siteng Huang, Min Zhang, Han Zhao, and Donglin Wang. 2024{\natexlab{c}}.
\newblock Pite: Pixel-temporal alignment for large video-language model.
\newblock In \emph{European Conference on Computer Vision}, pages 160--176. Springer.

\bibitem[{Loshchilov and Hutter(2019)}]{adamw}
Ilya Loshchilov and Frank Hutter. 2019.
\newblock Decoupled weight decay regularization.
\newblock In \emph{ICLR}.

\bibitem[{Maaz et~al.(2024{\natexlab{a}})Maaz, Rasheed, Khan, and Khan}]{videogpt+}
Muhammad Maaz, Hanoona Rasheed, Salman Khan, and Fahad Khan. 2024{\natexlab{a}}.
\newblock Videogpt+: Integrating image and video encoders for enhanced video understanding.
\newblock \emph{arXiv preprint arXiv:2406.09418}.

\bibitem[{Maaz et~al.(2024{\natexlab{b}})Maaz, Rasheed, Khan, and Khan}]{video-chatgpt}
Muhammad Maaz, Hanoona Rasheed, Salman Khan, and Fahad~Shahbaz Khan. 2024{\natexlab{b}}.
\newblock Video-chatgpt: Towards detailed video understanding via large vision and language models.
\newblock In \emph{ACL}.

\bibitem[{Microsoft.(2024)}]{phi3v}
Microsoft. 2024.
\newblock Phi-3.5 vision instruct.
\newblock \url{https://huggingface.co/microsoft/Phi-3.5-vision-instruct}.

\bibitem[{Moon et~al.(2023)Moon, Hyun, Park, Park, and Heo}]{moon2023querydependentvideorepresentationmoment}
WonJun Moon, Sangeek Hyun, SangUk Park, Dongchan Park, and Jae-Pil Heo. 2023.
\newblock Query-dependent video representation for moment retrieval and highlight detection.
\newblock In \emph{Proceedings of the IEEE/CVF Conference on Computer Vision and Pattern Recognition}, pages 23023--23033.

\bibitem[{Munasinghe et~al.(2024)Munasinghe, Gani, Zhu, Cao, Xing, Khan, and Khan}]{munasinghe2024videoglamm}
Shehan Munasinghe, Hanan Gani, Wenqi Zhu, Jiale Cao, Eric Xing, Fahad~Shahbaz Khan, and Salman Khan. 2024.
\newblock Videoglamm: A large multimodal model for pixel-level visual grounding in videos.
\newblock \emph{arXiv preprint arXiv: 2411.04923}.

\bibitem[{Peng et~al.(2024)Peng, Meng, Chen, Xie, Liu, Gui, Xu, Qiu, Wu, and Jiang}]{peng2024instit}
Wujian Peng, Lingchen Meng, Yitong Chen, Yiweng Xie, Yang Liu, Tao Gui, Hang Xu, Xipeng Qiu, Zuxuan Wu, and Yu-Gang Jiang. 2024.
\newblock Inst-it: Boosting multimodal instance understanding via explicit visual prompt instruction tuning.
\newblock \emph{arXiv preprint arXiv: 2412.03565}.

\bibitem[{Qian et~al.(2024)Qian, Li, Wu, Ye, Fei, Chua, Zhuang, and Tang}]{momentor}
Long Qian, Juncheng Li, Yu~Wu, Yaobo Ye, Hao Fei, Tat-Seng Chua, Yueting Zhuang, and Siliang Tang. 2024.
\newblock Momentor: Advancing video large language model with fine-grained temporal reasoning.
\newblock In \emph{ICML}.

\bibitem[{Radford et~al.(2021)Radford, Kim, Hallacy, Ramesh, Goh, Agarwal, Sastry, Askell, Mishkin, Clark et~al.}]{clip}
Alec Radford, Jong~Wook Kim, Chris Hallacy, Aditya Ramesh, Gabriel Goh, Sandhini Agarwal, Girish Sastry, Amanda Askell, Pamela Mishkin, Jack Clark, et~al. 2021.
\newblock Learning transferable visual models from natural language supervision.
\newblock In \emph{International conference on machine learning}, pages 8748--8763. PMLR.

\bibitem[{Ren et~al.(2024)Ren, Yao, Li, Sun, and Hou}]{timechat}
Shuhuai Ren, Linli Yao, Shicheng Li, Xu~Sun, and Lu~Hou. 2024.
\newblock Timechat: A time-sensitive multimodal large language model for long video understanding.
\newblock In \emph{CVPR}.

\bibitem[{Song et~al.(2024)Song, Chai, Wang, Zhang, Zhou, Wu, Chi, Guo, Ye, Zhang et~al.}]{song2024moviechat}
Enxin Song, Wenhao Chai, Guanhong Wang, Yucheng Zhang, Haoyang Zhou, Feiyang Wu, Haozhe Chi, Xun Guo, Tian Ye, Yanting Zhang, et~al. 2024.
\newblock Moviechat: From dense token to sparse memory for long video understanding.
\newblock In \emph{Proceedings of the IEEE/CVF Conference on Computer Vision and Pattern Recognition}, pages 18221--18232.

\bibitem[{Wang et~al.(2024{\natexlab{a}})Wang, Xu, Cheng, Diao, Zhou, Cao, Wang, Ge, and Huang}]{wang2024groundedvideollm}
Haibo Wang, Zhiyang Xu, Yu~Cheng, Shizhe Diao, Yufan Zhou, Yixin Cao, Qifan Wang, Weifeng Ge, and Lifu Huang. 2024{\natexlab{a}}.
\newblock Grounded-videollm: Sharpening fine-grained temporal grounding in video large language models.
\newblock \emph{arXiv preprint arXiv: 2410.03290}.

\bibitem[{Wang et~al.(2024{\natexlab{b}})Wang, Wang, Ye, Nie, and Huang}]{wang2024elysium}
Hang Wang, Yanjie Wang, Yongjie Ye, Yuxiang Nie, and Can Huang. 2024{\natexlab{b}}.
\newblock Elysium: Exploring object-level perception in videos via mllm.
\newblock \emph{ECCV}.

\bibitem[{Wang et~al.(2024{\natexlab{c}})Wang, Bai, Tan, Wang, Fan, Bai, Chen, Liu, Wang, Ge, Fan, Dang, Du, Ren, Men, Liu, Zhou, Zhou, and Lin}]{wang2024qwen2vl}
Peng Wang, Shuai Bai, Sinan Tan, Shijie Wang, Zhihao Fan, Jinze Bai, Keqin Chen, Xuejing Liu, Jialin Wang, Wenbin Ge, Yang Fan, Kai Dang, Mengfei Du, Xuancheng Ren, Rui Men, Dayiheng Liu, Chang Zhou, Jingren Zhou, and Junyang Lin. 2024{\natexlab{c}}.
\newblock Qwen2-vl: Enhancing vision-language model's perception of the world at any resolution.
\newblock \emph{arXiv preprint arXiv: 2409.12191}.

\bibitem[{Wang et~al.(2024{\natexlab{d}})Wang, Meng, Liang, Wang, Liu, and Zhao}]{hawkeye}
Yueqian Wang, Xiaojun Meng, Jianxin Liang, Yuxuan Wang, Qun Liu, and Dongyan Zhao. 2024{\natexlab{d}}.
\newblock Hawkeye: Training video-text llms for grounding text in videos.
\newblock \emph{arXiv preprint arXiv:2403.10228}.

\bibitem[{Wu et~al.(2024)Wu, Li, Chen, and Li}]{wu2024longvideobench}
Haoning Wu, Dongxu Li, Bei Chen, and Junnan Li. 2024.
\newblock \href {https://openreview.net/forum?id=3G1ZDXOI4f} {Longvideobench: A benchmark for long-context interleaved video-language understanding}.
\newblock In \emph{The Thirty-eight Conference on Neural Information Processing Systems Datasets and Benchmarks Track}.

\bibitem[{Wu et~al.(2025)Wu, Zhao, Li, Li, Zhou, Shou, and Bai}]{wu2025large}
Weijia Wu, Yuzhong Zhao, Zhuang Li, Jiahong Li, Hong Zhou, Mike~Zheng Shou, and Xiang Bai. 2025.
\newblock A large cross-modal video retrieval dataset with reading comprehension.
\newblock \emph{Pattern Recognition}, 157:110818.

\bibitem[{Xu et~al.(2024)Xu, Zhao, Zhou, Lin, Ng, and Feng}]{pllava}
Lin Xu, Yilin Zhao, Daquan Zhou, Zhijie Lin, See~Kiong Ng, and Jiashi Feng. 2024.
\newblock Pllava: Parameter-free llava extension from images to videos for video dense captioning.
\newblock \emph{arXiv preprint arXiv:2404.16994}.

\bibitem[{Yan et~al.(2023)Yan, Xiong, Nagrani, Arnab, Wang, Ge, Ross, and Schmid}]{yan2023unlocunifiedframeworkvideo}
Shen Yan, Xuehan Xiong, Arsha Nagrani, Anurag Arnab, Zhonghao Wang, Weina Ge, David Ross, and Cordelia Schmid. 2023.
\newblock Unloc: A unified framework for video localization tasks.
\newblock In \emph{Proceedings of the IEEE/CVF International Conference on Computer Vision}, pages 13623--13633.

\bibitem[{Yang et~al.(2023)Yang, Nagrani, Seo, Miech, Pont-Tuset, Laptev, Sivic, and Schmid}]{yang2023vid2seqlargescalepretrainingvisual}
Antoine Yang, Arsha Nagrani, Paul~Hongsuck Seo, Antoine Miech, Jordi Pont-Tuset, Ivan Laptev, Josef Sivic, and Cordelia Schmid. 2023.
\newblock Vid2seq: Large-scale pretraining of a visual language model for dense video captioning.
\newblock In \emph{CVPR}, pages 10714--10726.

\bibitem[{Yang et~al.(2024)Yang, Huang, Chai, Jiang, and Hwang}]{yang2024samurai}
Cheng-Yen Yang, Hsiang-Wei Huang, Wenhao Chai, Zhongyu Jiang, and Jenq-Neng Hwang. 2024.
\newblock Samurai: Adapting segment anything model for zero-shot visual tracking with motion-aware memory.
\newblock \emph{arXiv preprint arXiv: 2411.11922}.

\bibitem[{Yuan et~al.(2025)Yuan, Li, Zhang, Huang, Xu, Ji, Tong, Qi, Feng, and Yang}]{yuan2025sa2va}
Haobo Yuan, Xiangtai Li, Tao Zhang, Zilong Huang, Shilin Xu, Shunping Ji, Yunhai Tong, Lu~Qi, Jiashi Feng, and Ming-Hsuan Yang. 2025.
\newblock Sa2va: Marrying sam2 with llava for dense grounded understanding of images and videos.
\newblock \emph{arXiv preprint arXiv: 2501.04001}.

\bibitem[{Yuan et~al.(2024)Yuan, Zhang, Li, Cheng, Zhang, Li, Li, Zhao, Zhang, Zhuang, Zhu, and Bing}]{yuan2024videorefer}
Yuqian Yuan, Hang Zhang, Wentong Li, Zesen Cheng, Boqiang Zhang, Long Li, Xin Li, Deli Zhao, Wenqiao Zhang, Yueting Zhuang, Jianke Zhu, and Lidong Bing. 2024.
\newblock Videorefer suite: Advancing spatial-temporal object understanding with video llm.
\newblock \emph{arXiv preprint arXiv: 2501.00599}.

\bibitem[{Zeng et~al.(2024{\natexlab{a}})Zeng, Li, Wang, Li, Jiang, Yan, Li, Shi, Yue, Wang, Wang, Qiao, and Wang}]{zeng2024timesuite}
Xiangyu Zeng, Kunchang Li, Chenting Wang, Xinhao Li, Tianxiang Jiang, Ziang Yan, Songze Li, Yansong Shi, Zhengrong Yue, Yi~Wang, Yali Wang, Yu~Qiao, and Limin Wang. 2024{\natexlab{a}}.
\newblock Timesuite: Improving mllms for long video understanding via grounded tuning.
\newblock \emph{arXiv preprint arXiv: 2410.19702}.

\bibitem[{Zeng et~al.(2024{\natexlab{b}})Zeng, Li, Wang, Li, Jiang, Yan, Li, Shi, Yue, Wang et~al.}]{timesuite}
Xiangyu Zeng, Kunchang Li, Chenting Wang, Xinhao Li, Tianxiang Jiang, Ziang Yan, Songze Li, Yansong Shi, Zhengrong Yue, Yi~Wang, et~al. 2024{\natexlab{b}}.
\newblock Timesuite: Improving mllms for long video understanding via grounded tuning.
\newblock \emph{arXiv preprint arXiv:2410.19702}.

\bibitem[{Zhang et~al.(2023)Zhang, Li, and Bing}]{video-llama}
Hang Zhang, Xin Li, and Lidong Bing. 2023.
\newblock Video-llama: An instruction-tuned audio-visual language model for video understanding.
\newblock \emph{arXiv preprint arXiv:2306.02858}.

\bibitem[{Zhang et~al.(2024{\natexlab{a}})Zhang, Li, Liu, Lee, Gui, Fu, Feng, Liu, and Li}]{llava2024}
Yuanhan Zhang, Bo~Li, Haotian Liu, Yong~Jae Lee, Liangke Gui, Di~Fu, Jiashi Feng, Ziwei Liu, and Chunyuan Li. 2024{\natexlab{a}}.
\newblock Llava next video.
\newblock \url{https://llava-vl.github.io/blog/2024-04-30-llava-next-video/}.

\bibitem[{Zhang et~al.(2024{\natexlab{b}})Zhang, Wu, Li, Li, Ma, Liu, and Li}]{zhang2024llava-video}
Yuanhan Zhang, Jinming Wu, Wei Li, Bo~Li, Zejun Ma, Ziwei Liu, and Chunyuan Li. 2024{\natexlab{b}}.
\newblock Video instruction tuning with synthetic data.
\newblock \emph{arXiv preprint arXiv: 2410.02713}.

\end{thebibliography}
\clearpage
\appendix

\section{More Details of \datasetname}
\label{sec:more_dataset_details}

\subsection{Data Statics of \datasetname}
\label{data_statics}

\begin{figure}
    \centering
    \subfigure{
        \includegraphics[width=0.45\textwidth]{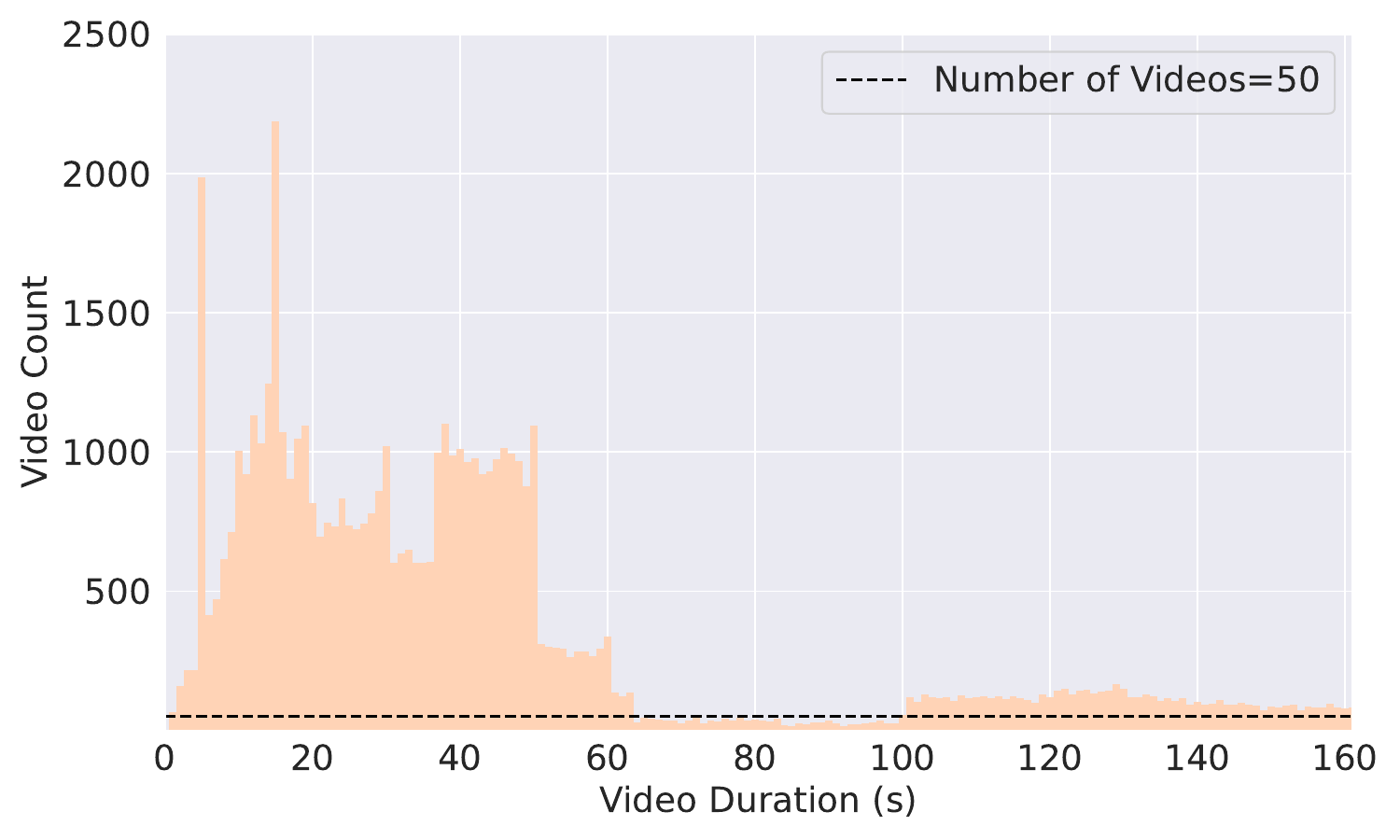}
        \label{fig:subfig1}
    }
    \hfill
    \subfigure{
        \includegraphics[width=0.45\textwidth]{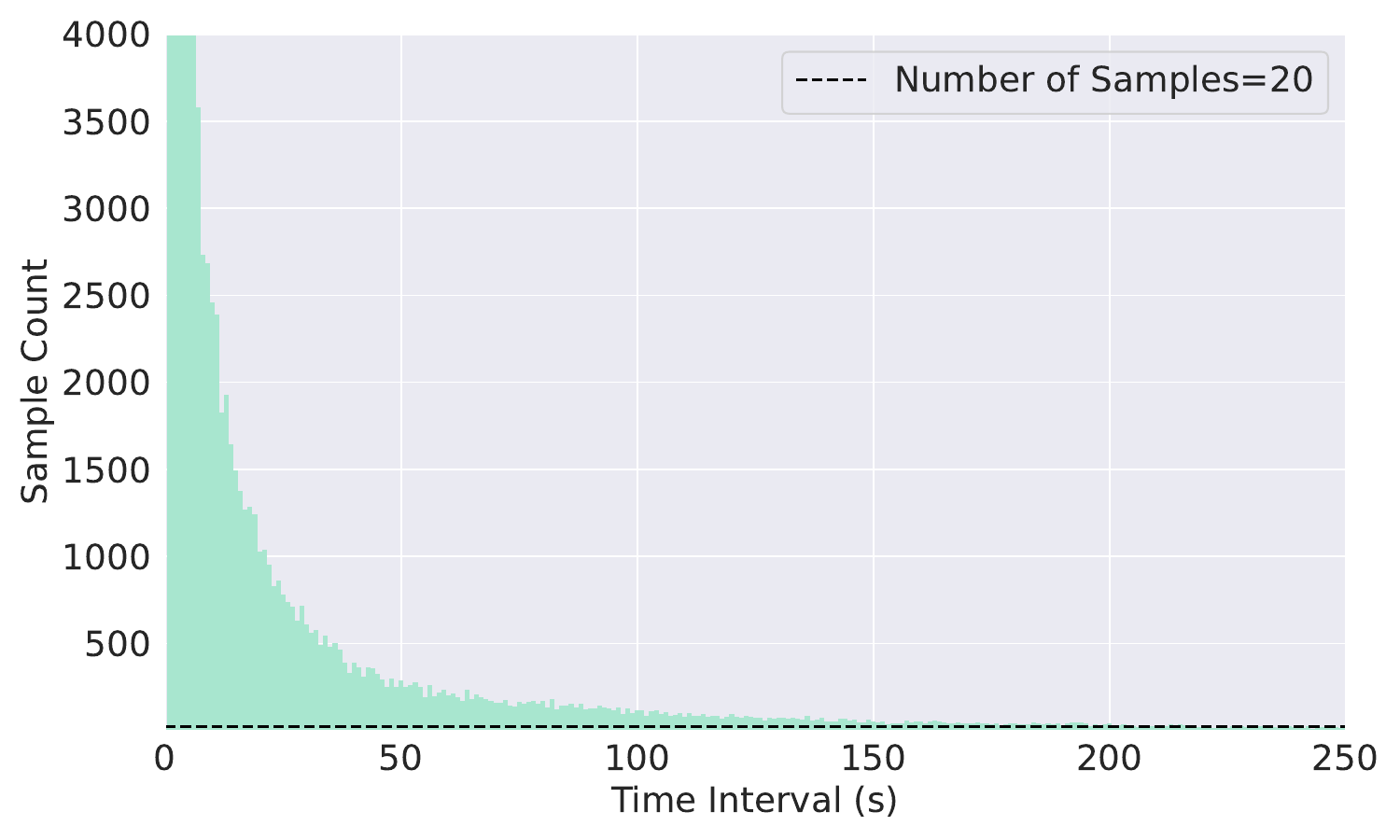}
        \label{fig:subfig2}
    }    
    \caption{(a) Video duration visualization for iMOVE-IT. (b) Visualization of time intervals in iMOVE-IT.}
    \label{fig:subfigs}
\end{figure}

\paragraph{Diversity of Video Types}
The iMOVE-IT dataset contains 68387 unique videos from 11 different datasets, including InternVid-10M, CLEVRER, DiDeMo, NExT-QA, HACS, YT-Temporal-1B, COIN, TACoS, YouCook2, HiREST and ViTT, and does not include videos from the ActivityNet and Charades-STA datasets. The average duration of the videos is 109 seconds. The distribution of video durations in iMOVE-IT is shown in Figure ~\ref{fig:subfig1}. As can be seen from the figure, the distribution of video durations is quite broad, with the majority of videos being less than 60 seconds long. Videos longer than 60 seconds are primarily concentrated in the range of 100 to 160 seconds.

\paragraph{Diversity of Object Types} 
The iMOVE-IT dataset encompasses 114,705 objects, derived from 23,278 object categories, with an average of 5 objects per category. Specifically, the Spatial grounding task consists of 51,992 objects, the Temporal Grounding task includes 21,328 objects, and the Instance Dynamic Captioning task comprises 41,385 objects.

\paragraph{Diversity of time intervals} 
Figure ~\ref{fig:subfig2} depicts the distribution of time intervals in iMOVE-IT, with the majority of time intervals concentrated within 50 seconds. This is likely due to the fact that the video durations are primarily between 1 and 60 seconds. Additionally, the diversity in the distribution of time intervals contributes to enhancing the model's robustness.

From the above analysis, we can see that the iMOVE-IT dataset has significant advantages in terms of sample size, video types, object types, time intervals, and task instruction diversity. These characteristics make this dataset highly valuable for research and applications in related fields.

\clearpage
\subsection{Task Prompts of \datasetname}
\label{task_prompt}

\begin{center}
\begin{tcolorbox}[colback=gray!20, colframe=black, text width=0.9\textwidth, title={Prompts of iMOVE-IT}]
\textcolor{blue}{Spatial Grounding} \\
1. Please give the bounding box coordinates variation of the object depicted as \verb|<dynamic caption>| during the time interval from \verb|<t1>| to \verb|<t2>|. Output only the bounding box coordinates for the start and end times. \\
2. I would like to know the bounding box coordinates variation of the object described as \verb|<dynamic caption>| during the period from \verb|<t1>| to \verb|<t2>|. Just produce the bounding box coordinates for when it starts and ends. \\
3. Can you give me the bounding box coordinates change for the object depicted as \verb|<dynamic caption>| from \verb|<t1>| to \verb|<t2>|? Only the bounding box coordinates for the beginning and conclusion should be generated. \\
4. Kindly provide the variation in the bounding box coordinates of the object described as \verb|<dynamic caption>| from \verb|<t1>| to \verb|<t2>|. Simply output the bounding box coordinates that mark the start and end. \\
5. What is the change in the bounding box coordinates of the object depicted as \verb|<dynamic caption>| during the interval from \verb|<t1>| to \verb|<t2>|? Provide solely the bounding box coordinates that denote the start and end points. \\
6. Let me know the variation in the bounding box coordinates of the object described as \verb|<dynamic caption>| during the time span from \verb|<t1>| to \verb|<t2>|. Ensure the output includes only the bounding box coordinates for the start and end. \\
7. I need the bounding box coordinates change of the object depicted as \verb|<dynamic caption>| during the period from \verb|<t1>| to \verb|<t2>|. Limit the output to the bounding box coordinates of the start and end times. \\
8. Kindly tell me the change in the bounding box coordinates for the object depicted as \verb|<dynamic caption>| between \verb|<t1>| and \verb|<t2>|. You need to output solely the bounding box coordinates for the start and end times. \\

\textcolor{blue}{Temporal Grounding} \\
1. Please give the time interval when the object described as \verb|<dynamic caption>| transitions from the bounding box coordinates \verb|<bbox1>| to \verb|<bbox2>|. \\
2. Could you provide the time interval for the transition of the object, described as \verb|<dynamic caption>|, from the bounding box coordinates \verb|<bbox1>| to \verb|<bbox2>|? \\
3. What is the time interval when the object, depicted as \verb|<dynamic caption>|, shifts from the bounding box \verb|<bbox1>| to \verb|<bbox2>|? \\
4. Kindly indicate the time span for the object, described as \verb|<dynamic caption>|, transitioning from \verb|<bbox1>| to \verb|<bbox2>|. \\
5. Please let me know the duration of time when the object, with a description of \verb|<dynamic caption>|, moves from the coordinates \verb|<bbox1>| to \verb|<bbox2>|. \\
6. Identify the time interval for the object, described as \verb|<dynamic caption>|, as it transitions from \verb|<bbox1>| to \verb|<bbox2>|. \\
7. Could you specify the time period for the object, with a depiction of \verb|<dynamic caption>|, moving from \verb|<bbox1>| to \verb|<bbox2>|? \\
8. Please tell me the duration of time for the object, with a  portrayal of \verb|<dynamic caption>|, as it transitions from \verb|<bbox1>| to \verb|<bbox2>|. \\
\end{tcolorbox}
\end{center}
\clearpage

\begin{center}
\begin{tcolorbox}[colback=gray!20, colframe=black, text width=0.9\textwidth, title={Prompts of iMOVE-IT (continued)}]
\textcolor{blue}{Instance Dynamic Captioning} \\ 
1. Please describe the object within the time interval from \verb|<t1>| to \verb|<t2>|, with bounding box coordinates starting at \verb|<bbox1>| and ending at \verb|<bbox2>|. Include the object's appearance, changes, and behavior. \\
2. Could you describe the object within the time interval from \verb|<t1>| to \verb|<t2>|, with bounding box coordinates beginning at \verb|<bbox1>| and ending at \verb|<bbox2>|, and include its appearance, changes, and behavior? \\
3. Describe the object from \verb|<t1>| to \verb|<t2>|, with bounding box coordinates that begin at \verb|<bbox1>| and end at \verb|<bbox2>|, and include information on its appearance, alterations, and behavior. \\
4. Could you give a description of the object that is positioned between \verb|<t1>| and \verb|<t2>|, with bounding box coordinates commencing at \verb|<bbox1>| and finishing at \verb|<bbox2>|, and include its appearance, changes, and behavior? \\
5. Would you describe the object in the time span from \verb|<t1>| to \verb|<t2>|, with bounding box coordinates that begin at \verb|<bbox1>| and end at \verb|<bbox2>|, and mention its appearance, changes, and behavior? \\
6. Please provide a description of the object between \verb|<t1>| and \verb|<t2>|, with bounding box coordinates starting at \verb|<bbox1>| and ending at \verb|<bbox2>|, and include its appearance, changes, and behavior. \\
7. Describe the object from \verb|<t1>| to \verb|<t2>|, with bounding box coordinates initiating at \verb|<bbox1>| and concluding at \verb|<bbox2>|, and incorporate its appearance, variations, and behavior. \\
8. Could you portray the object from \verb|<t1>| to \verb|<t2>|, with bounding box coordinates that \break commence at \verb|<bbox1>| and finish at \verb|<bbox2>|, and include its appearance, changes, and \break behavior?
\end{tcolorbox}
\end{center}
\clearpage

\section{EXPERIMENTS}

\subsection{Detailed Dxperiment Setup}
\label{experiment_setup}
\begin{center}
\begin{table}[htbp]

\vskip 0.05in
\begin{tabular}{@{}lc@{}}
\toprule
Learning Rate                  & 1e-4             \\ 
LR Scheduler         & Cosine             \\
Global Batch Size              & 128              \\
Training Steps                 & 10K              \\
Warmup Ratio                   & 0.03                          \\
Trainable Modules              &     MLP Projector \& LoRA \\ 
Frame Resolution               & 448$\times$448               \\
Num Frames                & 96               \\ 
Train Epochs                  & 1 \\
Model Max Length                & 10000    \\
ZeRO Optimization              & Zero-2                       \\
Computation & 16 A800 \\
\bottomrule
\end{tabular}%
\caption{The hyper-parameters for iMOVE.}
\label{tab:training_settings}
\end{table}
\end{center}

We utilize InternViT-300M-448px~\cite{chen2024internvl} as the video encoder and Phi-3-mini-128k-instruct~\cite{abdin2024phi} as the LLM decoder. The parameters of equivalent components from InternVL2-4B, including linear projection layers, are used to initialize these models. The maximum value for temporal token quantization, denoted as \( Z \), is set to 300. For spatial tokens, the maximum quantization values for height \( \hat{H} \) and width \( \hat{W} \) are both set to 1000. We fine-tune the LLM using LoRA~\cite{lora}, while keeping the visual encoder frozen and making the MLP trainable. The LoRA parameters are set to \( r = 128 \) and \( \alpha = 256 \). We employ the AdamW~\cite{adamw} optimizer with a warm-up rate of 0.03. Initially, the video is divided into 96 segments, and one frame is randomly selected from each segment, resulting in 96 video frames. For each video, we select \( K = 24 \) events, setting the first frame feature's \( h \) and \( w \) of each event to 8, thus encoding with 64 tokens. The stride \( s \) is set to 2, meaning that non-first frames in each event are encoded with 16 tokens, resulting in a total of 2688 tokens per video, which is significantly fewer than the 4096 tokens per video encoding method in InternVL2-4B. Additional hyper-parameters can be found in Table~\ref{tab:training_settings}. All experiments are conducted on 16 A800 GPUs with a batch size of 128. The training set we used contains 1,276,977 data samples, and the experiment took a total of 41 hours to complete. Furthermore, in the construction of iMOVE-IT, we utilized the ViT-B/32 version of CLIP.

\begin{table*}

\resizebox{\textwidth}{!}{%
\begin{tabular}{@{}c|c|c@{}}
\toprule
\textbf{Task} & \textbf{\# of Entries} & \textbf{Datasets} \\ 
\midrule
\midrule
Temporal Sentence Grounding & 194K & DiDeMo, HiREST, QuerYD, VTG-IT-MR, TACOS \\
Temporal Action Localization & 45K & HACS \\
Dense Video Captioning & 61K & COIN, ViTT, YouCook2 \\
VideoQA & 493K & EgoQA, NExT-QA, Intent-QA, CLEVRER, LLAVA-Video-QA \\
Classification & 66K & SthSthV2, Kinetics \\
Video Captioning & 276K & YouCook2, WebVid-stage3, LLAVA-Video-cap \\
iMOVE-IT & 114K & self-collected \\ 
\bottomrule
\end{tabular}%
}
\caption{Dataset Composition.}
\label{tab:train_datasets}
\end{table*}

\begin{table*}

\resizebox{\textwidth}{!}{%
\begin{tabular}{@{}c|c|c@{}}
\toprule
\textbf{Task} & \textbf{\# of Entries} & \textbf{Datasets} \\ 
\midrule
\midrule
Temporal Sentence Grounding & 194K & DiDeMo, HiREST, QuerYD, VTG-IT-MR, TACOS \\
Temporal Action Localization & 45K & HACS \\
Dense Video Captioning & 61K & COIN, ViTT, YouCook2 \\
\bottomrule
\end{tabular}%
}
\caption{Dataset Composition of the ablation study on Event-aware Spatiotemporal Efficient Modeling.}
\label{tab:train_datasets_abla_study}
\end{table*}

\subsection{Baseline and Comparison}
\label{appendix:baseline}
For the baseline models, the comparison primarily involves two categories: the first is the zero-shot video large language models, and the second is the expert models obtained through fine-tuning. For the first type of baseline method, we choose InternVL2-4B\citep{InternVL2024}, Video-ChatGPT~\cite{video-chatgpt}, VideoChat~\cite{videochat}, Momentor~\cite{momentor}, TimeChat~\cite{timechat}, VTG-LLM~\cite{vtgllm}, HawkEye~\cite{hawkeye}, PiTe \citep{pite}, Grounded-VideoLLM \citep{wang2024groundedvideollm}, VTimeLLM~\cite{vtimellm}, TimeSuite~\cite{zeng2024timesuite}, and TRACE~\cite{trace}. We selected Vid2Seq\citep{yang2023vid2seqlargescalepretrainingvisual} as the classic supervised expert model on the Charades-STA dataset. For the ActivityNet-Captions, we chose QD-DETR\citep{moon2023querydependentvideorepresentationmoment} and UnLoc-L\citep{yan2023unlocunifiedframeworkvideo} as the supervised expert models. Additionally, we report the results of some zero-shot models after fine-tuning.

\subsection{Detailed Data Filtering and Dataset Composition}\label{data_filtering}

In terms of data quality control, we implemented stringent measures. We excluded datasets such as STAR, ANet-RTL, VCG-Plus112K, Videochatgpt-100K, Videochat2-Conv, and TextVR to ensure a strict zero-shot setting on the Charades-STA and ActivityNet-Captions datasets. Additionally, the LLAVA-Video-cap and LLAVA-Video-QA datasets we used are subsets extracted from LLaVA-Video-178K, excluding Charades, ActivityNet, and Ego4D video sources, thus preventing potential data leakage. The data we utilized includes tasks such as Temporal Sentence Grounding, Temporal Action Localization, Dense Video Captioning, VideoQA, Classification, Video Captioning, and iMOVE-IT, with the detailed composition of each task shown in Table~\ref{tab:train_datasets}.

\subsection{Detailed Benchmarks and Evaluation Metrics}
\label{benchmark}
iMOVE is comprehensively evaluated across the following four tasks:

Temporal Video Grounding: This task aims to determine the temporal boundaries of a single event based on a text description. The datasets \textbf{Charades-STA}~\citep{charades-sta} and \textbf{ActivityNet-Captions}~\citep{activitynet} are used for evaluation. For this task, we report the Intersection over Union (IoU) between the predicted timestamps by the model and the ground truth annotations. Specifically, we calculate \textbf{Recall at IoU} thresholds of \{0.3, 0.5, 0.7\} and their \textbf{mean IoU}.

Dense Video Captioning: This task is more complex, requiring the joint localization of key events and the generation of descriptions for each segment. The \textbf{ActivityNet-Captions} dataset is used for evaluation. We report \textbf{SODA\_c}~\cite{fujita2020soda}, which is specifically tailored for the video's storyline, and \textbf{METEOR}~\cite{banerjee2005meteor}, which is the average of traditional METEOR scores calculated based on matched pairs between generated events and the ground truth across IoU thresholds of \{0.3, 0.5, 0.7, 0.9\}.

General Video Understanding: This task aims to evaluate the general short-term video understanding capabilities of iMOVE. We utilize \textbf{Video-MME}~\cite{videomme} and \textbf{MVBench}~\cite{mvbench}, reporting their average accuracy.

Long-term Video Understanding: This task aims to evaluate the long-term video understanding capabilities of video models. We utilize the \textbf{LongVideoBench}~\cite{wu2024longvideobench} and the Long subset of \textbf{Video-MME}, reporting their average accuracy.

\subsection{Explanation of Methods on ActivityNet-Captions That Are Not Strictly Zero-Shot Setting
}\label{data_leakage}
Due to potential data leakage, some methods on ActivityNet-Captions do not strictly adhere to zero-shot settings. Below is a detailed explanation:

\textbf{PiTe} utilized the Video-ChatGPT~\cite{video-chatgpt} dataset in Stage 3, which was constructed using ActivityNet~\cite{activitynet} as the source. \par
\textbf{Grounded-VideoLLM} employed ANet-RTL, VCG-Plus-112K~\cite{videogpt+}, Videochatgpt-100K, and Videochat2-Conv in Stage 3, all of which were constructed using ActivityNet as the source. Additionally, the TextVR~\cite{wu2025large} used in this method also utilized videos from ActivityNet.  \par

\textbf{HawkEye} also utilizes VideoChatGPT and TextVR, resulting in non-strict zero-shot settings on the ActivityNet-Captions dataset. \par

\textbf{VTimeLLM} directly used the ActivityNet Captions~\cite{krishna2017dense} dataset as the training set in Stage 3.

\subsection{Data Composition of Ablation Study}\label{dataset_composition_abl}

 Due to computational resource considerations, we conducted the ablation experiment for Event-aware Spatiotemporal Efficient Modeling using only a subset of the entire dataset related to temporal understanding. As shown in Table~\ref{tab:train_datasets_abla_study}, this includes the tasks of Temporal Sentence Grounding, Temporal Action Localization, and Dense Video Captioning.

\section{Qualitative analysys}
\label{case_study}

\begin{figure}
    \centering
    \includegraphics[width=1.0\linewidth]{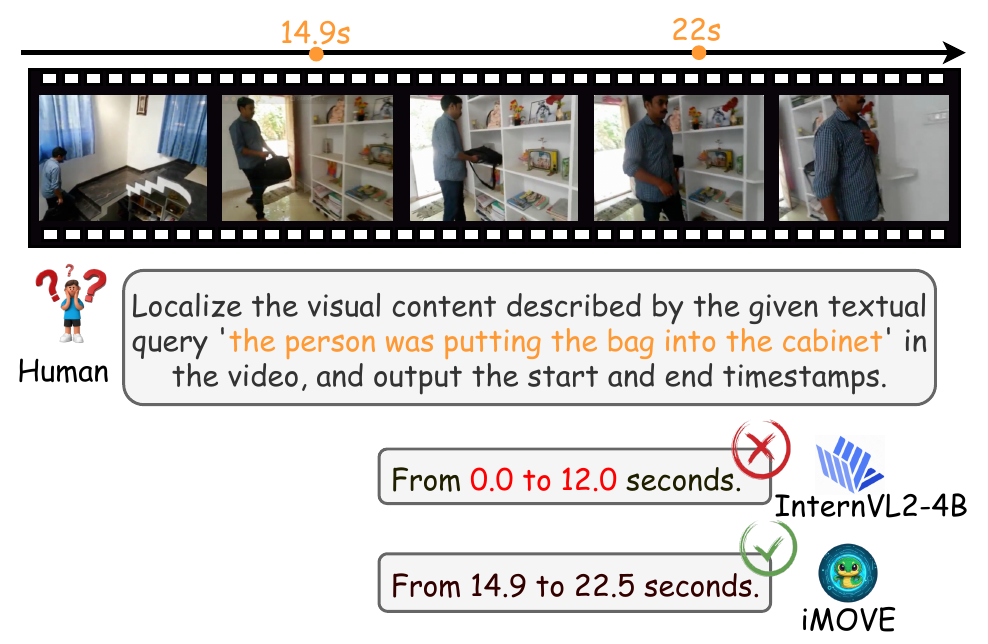}
    \caption{
     {Qualitative comparison of the temporal grounding capabilities of iMOVE and InternVL2-4B.}
    }
    \label{fig:qua_temporal_grounding}
    \vspace{-0.5cm}
\end{figure}

We provide a detailed qualitative comparison between iMOVE and InternVL2-4B in terms of temporal grounding, dense captioning, and long-term video understanding.

\textbf{Qualitative Comparison in Temporal Grounding.} As shown in Figure~\ref{fig:qua_temporal_grounding}, iMOVE accurately identifies the time interval of the event "the person was putting the bag into the cabinet" from a video containing multiple events, whereas InternVL2-4B fails to pinpoint the specific video segment where the event occurs. This demonstrates iMOVE's strong fine-grained temporal perception capability.

\begin{figure}
    \centering
    \includegraphics[width=1.0\linewidth]{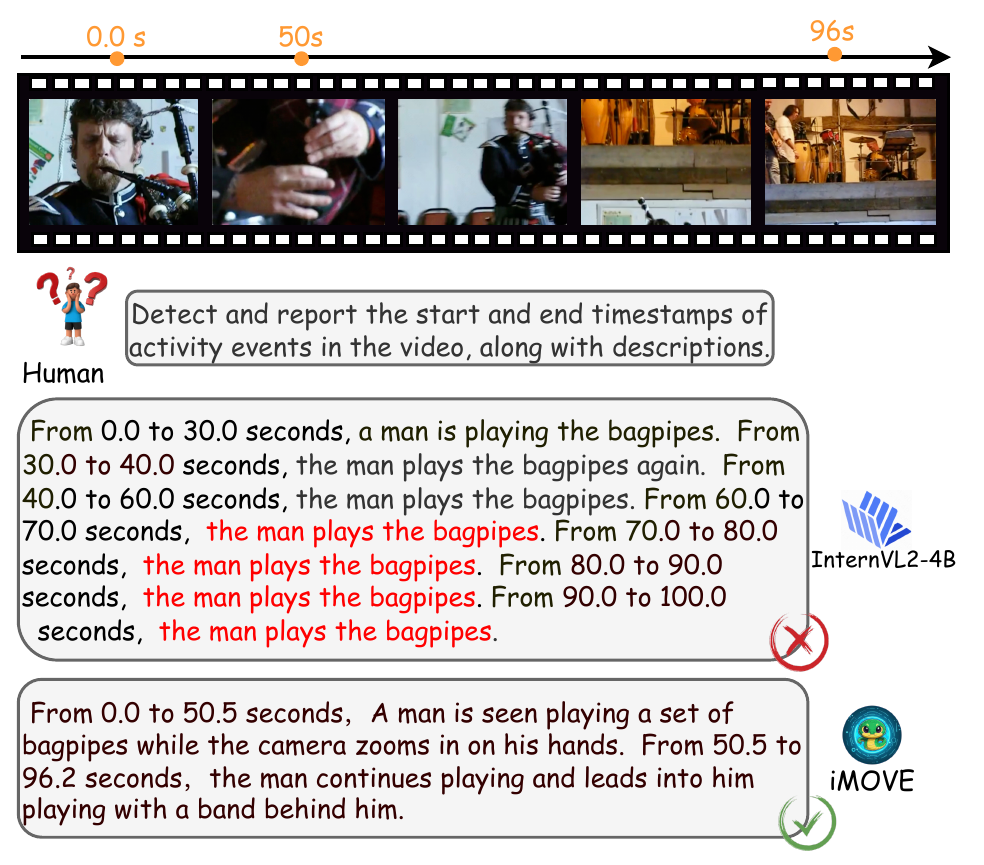}
    \caption{
     {Qualitative comparison of the dense captioning capabilities of iMOVE and InternVL2-4B.}
    }
    \label{fig:qua_dense_caption}
    \vspace{-0.5cm}
\end{figure}

\textbf{Qualitative Comparison in Dense Captioning.} As illustrated in Figure~\ref{fig:qua_dense_caption}, iMOVE effectively captures the complete storyline of the video, accurately identifying the time intervals of various events and providing precise event descriptions. In contrast, InternVL2-4B struggles to correctly comprehend multiple events within the video, resulting in repetitive event descriptions.

\begin{figure}
    \centering
    \includegraphics[width=1.0\linewidth]{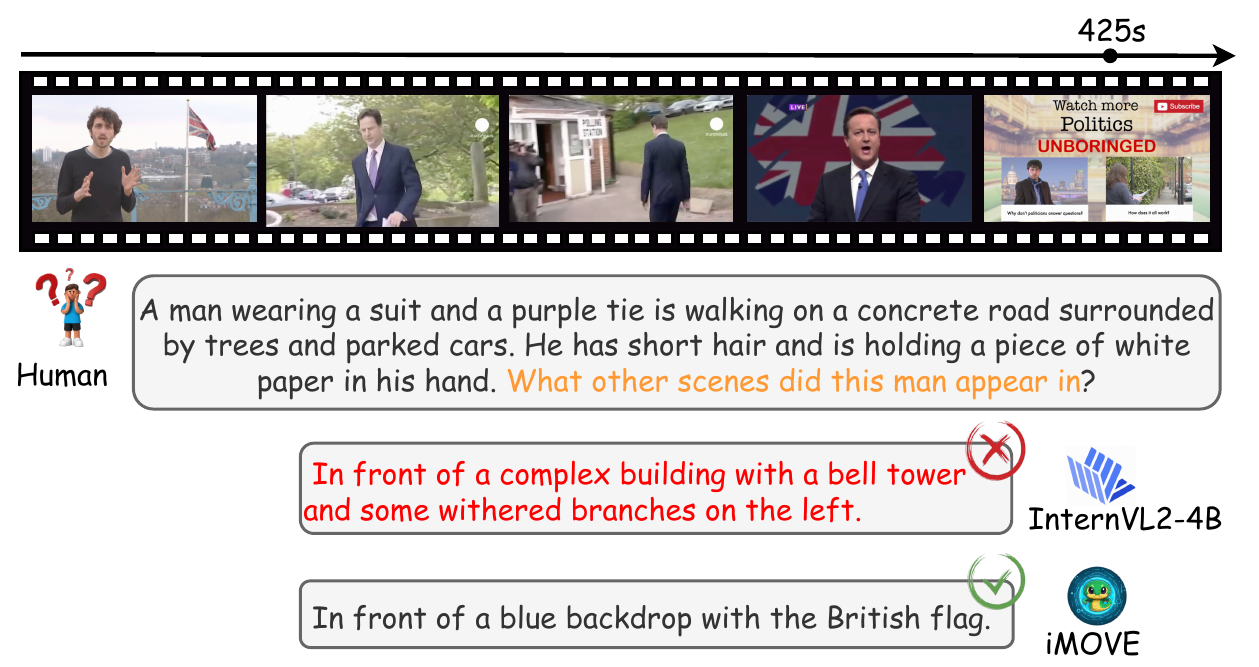}
    \caption{
     {Qualitative comparison of the long-term video understanding capabilities of iMOVE and InternVL2-4B.}
    }
    \label{fig:qua_long_video}
    \vspace{-0.5cm}
\end{figure}

\textbf{Qualitative Comparison in Long-term Video Understanding.} As depicted in Figure~\ref{fig:qua_long_video}, thanks to its meticulously designed architecture and dataset, iMOVE accurately answers reasoning questions in long videos. iMOVE first locates the segments described in the questions within the long video and determines the key characteristics of the man based on the video content, using them as crucial clues when he appears in another video segment. As a short video large language model, InternVL2-4B fails to provide correct answers.

In summary, iMOVE, with its carefully designed architecture based on iMOVE-IT, adapts a short video large language model to perceive fine-grained spatiotemporal instance motions, significantly enhancing its temporal understanding, general video understanding and long-term video understanding capabilities. This improvement in general video understanding offers a substantial advantage over previous temporal Video-LLMs, which excelled in temporal understanding but lacked in general video understanding capabilities.


\end{document}